\documentclass[10pt,twocolumn,letterpaper]{article}

\usepackage[pagenumbers]{iccv} 

\usepackage{graphicx}
\usepackage{amsmath}
\usepackage{amssymb}
\usepackage{booktabs}
\usepackage[dvipsnames]{xcolor}
\usepackage{comment}

\usepackage{wrapfig}
\usepackage{float}
\usepackage{tabularx}
\newcolumntype{Y}{>{\centering\arraybackslash}X}
\usepackage{epigraph}
\usepackage{scalerel}
\usepackage{arydshln}
\usepackage{stfloats} 

\usepackage[accsupp]{axessibility}  

\DeclareMathOperator*{\argmax}{arg\,max}

\newcommand{\squishlist}{
	\begin{list}{$\bullet$}
		{ \setlength{\itemsep}{0pt}
			\setlength{\parsep}{1pt}
			\setlength{\topsep}{1pt}
			\setlength{\partopsep}{0pt}
			\setlength{\leftmargin}{1.5em}
			\setlength{\labelwidth}{1em}
			\setlength{\labelsep}{0.5em} } }
\newcommand{\squishend}{\end{list} 
}

\definecolor{blue-violet}{rgb}{0.54, 0.17, 0.89}

\definecolor{iccvblue}{rgb}{0.21,0.49,0.74}
\usepackage[pagebackref,breaklinks,colorlinks,allcolors=iccvblue]{hyperref}

\def\paperID{41} 
\def\confName{ICCV}
\def\confYear{2025}

\title{
Robust 3D Object Detection using Probabilistic Point Clouds
\\from Single-Photon LiDARs
}

\author{
Bhavya Goyal\textsuperscript{1}\quad Felipe Gutierrez-Barragan\textsuperscript{2}\quad Wei Lin\textsuperscript{1}\quad Andreas Velten\textsuperscript{1,2}\quad Yin Li\textsuperscript{1}\quad Mohit Gupta\textsuperscript{1,2}\\
\textit{\textsuperscript{1}University of Wisconsin-Madison \qquad \textsuperscript{2}Ubicept}
}

\begin{document}
\maketitle

\begin{abstract}

\vspace{-5pt}

LiDAR-based 3D sensors provide point clouds, a canonical 3D representation used in various scene understanding tasks. Modern LiDARs face key challenges in several real-world scenarios, such as long-distance or low-albedo objects, producing sparse or erroneous point clouds. These errors, which are rooted in the noisy raw LiDAR measurements, get propagated to downstream perception models, resulting in potentially severe loss of accuracy. This is because conventional 3D processing pipelines do not retain any uncertainty information from the raw measurements when constructing point clouds.

We propose Probabilistic Point Clouds (PPC), a novel 3D scene representation where each point is augmented with a probability attribute that encapsulates the measurement uncertainty (or confidence) in the raw data. We further introduce inference approaches that leverage PPC for robust 3D object detection; these methods are versatile and can be used as computationally lightweight drop-in modules in 3D inference pipelines. We demonstrate, via both simulations and real captures, that PPC-based 3D inference methods outperform several baselines using LiDAR as well as camera-LiDAR fusion models, across challenging indoor and outdoor scenarios involving small, distant, and low-albedo objects, as well as strong ambient light.
Our project webpage is at \normalfont{
\href{https://bhavyagoyal.github.io/ppc}{https://bhavyagoyal.github.io/ppc}}.
\end{abstract}

\vspace{-5pt}
\section{Introduction}
\label{sec:intro}

\begin{figure}[htp!]
\centering
\vspace{-2pt}
\includegraphics[width=\linewidth]{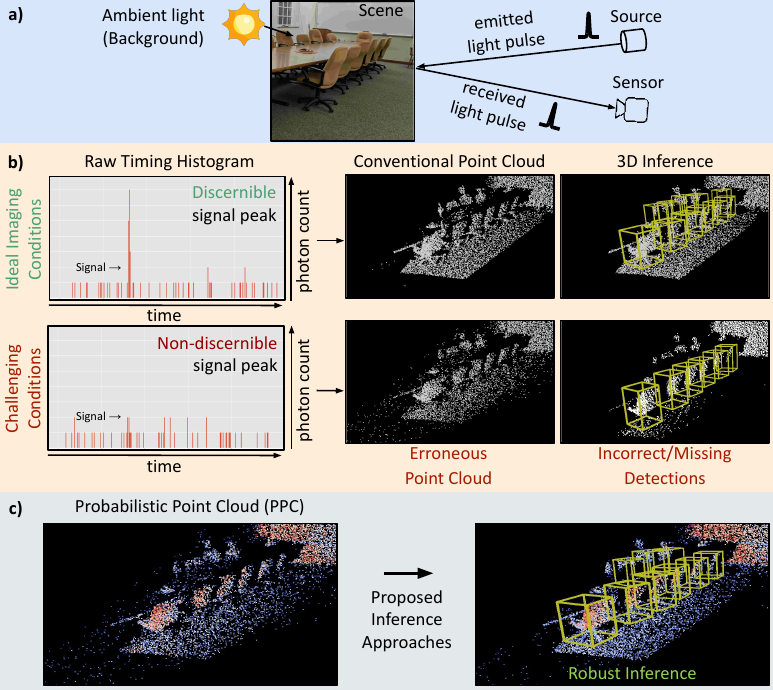}
\vspace{-12pt}
\caption{\textbf{Robust 3D Inference under Challenging Real-World Conditions:} a) A LiDAR provides depth by measuring time delays between emitted and received pulses. b) The raw sensor data at each LiDAR pixel is a temporal histogram of photon counts vs.\ time. The time location of the histogram peak corresponds to the scene distance, which is then converted to a point in the 3D point cloud.
Under challenging conditions, the signal peaks cannot be easily discerned from the background, resulting in large errors in the point clouds and incorrect inference. c) We propose Probabilistic Point Clouds (PPC), a scene representation with a probability attribute for each point. The figure shows a real LiDAR capture, where PPC is visualized with heatmap colors. Leveraging PPC, our proposed approach achieves robust 3D inference even under challenging scenarios.}
\vspace{-10pt}
\label{fig:noise_ppc}
\end{figure}

LiDARs are a prominent 3D imaging modality driving several applications, from embodied perception and autonomous vehicles~\cite{wang2024embodiedscan,li2020lidar}, to surveillance~\cite{Guo_2024_CVPR}, and more recently, deployed even in consumer devices (e.g., Apple iPhones). LiDARs are based on the time-of-flight (ToF) principle; a typical LiDAR consists of a laser source that emits short sub-nanosecond light pulses into the scene and a sensor that captures the reflected pulses; scene depths are estimated by computing the time delay between emitted and received pulses~\cite{ubsi_178} (Fig.~\ref{fig:noise_ppc}a). Increasingly, single photon avalanche diodes (SPADs)~\cite{cova1996avalanche} are becoming the sensor-of-choice in LiDARs due to their high sensitivity~\cite{pellegrini2000laser}, and amenability to fabrication of high-resolution arrays~\cite{morimoto20213}. It is not a surprise that the next generation of LiDAR devices is dominated by solid-state, high-resolution, and low-cost SPAD technology. 


A typical single-photon LiDAR builds a histogram of photon counts over time (Fig.~\ref{fig:noise_ppc}b). Under ideal imaging conditions, the peak in the timing histogram can be reliably detected, resulting in an accurate estimation of the time delay and hence the scene depth (Fig.~\ref{fig:noise_ppc}b). This estimated depth at each pixel can then be used to construct a 3D point cloud-based scene representation, a canonical input to various downstream 3D perception tasks~\cite{li2020lidar}.

However, under several real-world conditions, the raw timing histograms do not have a single, clearly discernible peak (Fig.~\ref{fig:noise_ppc}b). This could be due to a variety of factors, including a) distant objects or low scene albedo, b) strong ambient illumination which increases the background level and noise, c) other sources \eg non-diffuse and specular materials, multi-path or multi-camera interference, weather phenomena such as rain, snow, fog \cite{pellegrini2000laser,pediredla2018signal,beer2018background, satat2018towards} etc. Imagine a LiDAR mounted on an autonomous vehicle that needs to safely navigate the world, not just under fair lighting and weather, but across all operating conditions. Under these scenarios, it is often challenging to detect the correct peak location, resulting in large depth errors.

This challenge is typically addressed by filtering techniques that retain only the measurements with prominent peaks, while discarding the rest~\cite{zhang2013real,chen2020adaptive,goudreault2023lidar}. Therefore, in challenging scenarios mentioned above, most raw histograms that contain small or ambiguous peaks either get filtered out, resulting in incorrect removal of scene components, or introduce significant noise in the final point cloud if they are retained. Such filtering steps severely affect the inference performance by: (1) removing critical scene content in low signal regimes due to over-aggressive filtering, or (2) propagating excessive measurement noise to downstream inference models.\footnote{This filtering is a pre-processing step before outputting the final point cloud. Consequently, such noise is not observed in widely-used point cloud datasets as they consist of processed point clouds where low-confidence observations have already been removed, e.g., a distant pedestrian wearing dark clothing. Hence, a considerable amount of useful scene information is irretrievably lost from such point clouds.}

We present a different approach. Our key observation is that raw SPAD histograms encode rich scene information, which traditional LiDAR signal-processing approaches overlook by relying solely on peak locations to estimate point clouds.
Instead of \emph{deterministically} removing/retaining depth measurements, we propose augmenting point clouds with meaningful ``confidence'' features that encode physics-based information about the raw sensor measurements that could be valuable for downstream inference.
These confidence features 
require only lightweight compute operations. This is a critical consideration since these operations need to be performed on/close to the sensor, where compute and memory resources are extremely scarce. 

\vspace{-7pt}
\renewcommand {\epigraphflush}{center}
\setlength\epigraphwidth{0.96\linewidth}
\setlength\epigraphrule{0pt}
\epigraph{\small \emph{``Confidence. If you have it, you can make anything look good."}}{Diane Von Furstenberg}
\vspace{-7pt}

We use these confidence measures to create \emph{probabilistic point clouds (PPC)}, a novel 3D scene representation where each scene point is augmented with the confidence (or probability) attribute.
Going further, we leverage PPC representation to design computationally cheap inference approaches that are plug-and-play compatible with a wide range of existing 3D perception models, without needing to alter the architectures.
We demonstrate the effectiveness of our approach using widely employed point cloud-based 3D object detection models across both indoor and outdoor scenarios.
Despite its simplicity, the proposed approach achieves significant performance gains under challenging conditions, such as low-albedo or distant objects and intense ambient illumination.


This paper takes the first steps toward designing an \emph{end-to-end 3D inference pipeline} that is capable of propagating the uncertainty in depth measurements, starting from raw SPAD data, to downstream 3D inference models. 
Our contributions can be summarized as:

\squishlist
\item Designing lightweight and physically meaningful confidence features from raw SPAD histograms.
\item Developing a geometric scene representation (PPC) that propagates sensor uncertainty to point clouds.
\item Designing computationally low-cost approaches that utilize PPCs for robust 3D inference.
\item Demonstrating the performance of PPCs for 3D inference tasks using simulated and real LiDAR frames.
\squishend

\section{Related Work}
\noindent
\textbf{3D Inference using Point Clouds:} 
Point-based feature extraction networks \cite{qi2017pointnet, qi2017pointnet++} have become a standard building block for 3D inference~\cite{qi2019deep,shi2019pointrcnn,qi2018frustum,yang20203dssd,yin2021center}. 
These methods largely evaluate using clean point clouds benchmarks, but not on noisy point clouds; in fact, several works \cite{qi2017pointnet,zhang2024r,liu2020tanet} explicitly mention that network backbones are \emph{not robust} to noise in the point clouds.
Although a lot of work has been done for 2D image-based inference under challenging scenarios  \cite{gnanasambandam2020image,goyal2021photon,diamond2021dirty,hendrycks2019benchmarking, goyal2022robust}, there is surprisingly little prior work for 3D inference under physically-accurate sensor noise, especially for LiDARs. We show the performance degradation of widely used 3D detection models under real-world noise, and design approaches that perform robustly under such scenarios.

\medskip

\noindent \textbf{Confidence Attribute in 3D Inference:} While methods have been proposed to predict depth uncertainty for a camera image as part of monocular depth estimation \cite{bae2022multi,xia2020generating}, there are no prior works that leverage depth confidence or uncertainty for 3D LiDAR sensor data. This is largely because of limited access to raw sensor data.
Our work is a first in designing and propagating such confidence attributes for robust 3D inference.

\medskip

\noindent
\textbf{Point Cloud Denoising}: Denoising could potentially be used to reduce the noise in point clouds, and multiple algorithmic~\cite{digne2017bilateral,wolff2016point} and learning-based~\cite{BaoruiNoise2NoiseMapping,rakotosaona2020pointcleannet,luo2021score,hermosilla2019total,luo2020differentiable} solutions have been proposed.
These methods often introduce an extra, often significant, computation step to the 3D inference pipeline.
In this paper, we evaluate and compare the performance and computational cost of these denoising methods under a wide gamut of challenging conditions.


\medskip

\noindent
\textbf{3D Reconstruction from LiDAR:}
It has been shown that directly denoising raw LiDAR timing histograms can increase 3D reconstruction quality \cite{lindell2018single,peng2020photon,tachella2019real,rapp2017few, lee2023caspi}, albeit at the expense of extra computation. 
In this paper, we bypass the expensive denoising step and perform 3D inference on the lighter-weight, but potentially noisy, PPC representation extracted from the raw LiDAR data.
Furthermore, our approach is complementary to 3D reconstruction techniques. Hence, PPC can provide performance gains even with reconstructed point clouds. 


\section{Background: 3D Sensing Model for LiDAR}
A LiDAR imaging system typically consists of a synchronized pulsed laser source and a high-speed time-resolved detector such as a single-photon avalanche diode (SPAD)~\cite{cova1996avalanche}.
The laser source transmits a pulsed signal $s(t)$, \eg a Gaussian pulse with a repetition period of $\mathcal{T}_r$.
The scene is illuminated in a raster-scan manner or flood-illuminated to cover the detector's field of view, and the sensor observes the reflected light from the scene.
For each pixel location $(i,j)$, the incident photon flux $r_{i,j}$ is modelled~\cite{lindell2018single,peng2020photon} by the following equation: 
\vspace{-4pt}\begin{equation}
r_{i,j}[n] = \int_{n\Delta t}^{(n+1)\Delta t} \Phi_{i,j} \cdot s(t - \frac{2d_{i,j}}{\mathcal{C}})dt + b_\gamma,
\label{eq:photon_flux}
\end{equation}\vspace{-10pt}

\noindent where $\Delta t$ is the time duration of each discrete time-bin, $n \in \{1,2,...N\}$ where $N$ is the number of time bins over the duration of a laser period, $\Phi_{i,j}$ is a term that accounts for the distance fall-off, scene reflectance, and Bidirectional Reflectance Distribution Function (BRDF), $\mathcal{C}$ denotes the speed of light,
$s(t)$ denotes the pulsed signal from the laser, $d_{i,j}$ is the distance of the scene at the point of illumination $(i, j)$, and $b_\gamma$ denotes the photon flux from the ambient light.

Suppose the sensor has a quantum efficiency $\eta \in [0, 1)$, which describes the probability that the sensor can detect an incident photon. The detector's dark count is modeled as $b_d$, the number of spurious photon detections. Then, the number of photons measured by the sensor\footnote{Although we focus on SPAD histograms, the proposed analysis and techniques are compatible with other direct ToF time-resolved sensors, including avalanche photodiodes (APDs) and Silicon photomultipliers (SiPM), which follow a similar peak detection approach.} at each pixel can also be represented as a 1D timing histogram $h_{i,j}$ as follows:
\vspace{-4pt}\begin{equation}
h_{i,j}[n] \sim  \mathcal{P} \{ [\eta r_{i,j}[n] + b_d ] \}.
\label{eq:histogram}
\vspace{-2pt}
\end{equation}
\noindent The measurements are modeled as a Poisson process $\mathcal{P}$ with a time-varying arrival function $r_{i,j}[n]$ as the mean rate. Fig.~\ref{fig:noise_ppc} shows an example histogram captured under low and high noise. Distance estimate $\hat{d}_{i,j}$ for each location is computed using the bin with the highest photon count (peak location) from each timing histogram and by converting the time of flight to distance using the following equation:
\vspace{-4pt}\begin{equation}
\hat{d}_{i,j} = (\Delta t * \mathcal{C} /2) * \argmax_n\ h_{i,j}[n],
\label{eq:distance}
\vspace{-5pt}
\end{equation}
Finally, the distance estimates can be converted to coordinates in a 3D point cloud using camera intrinsic parameters.

\begin{figure*}[tp!]
\vspace{-5pt}

\begin{subfigure}{0.25\linewidth}
\includegraphics[width=\linewidth]{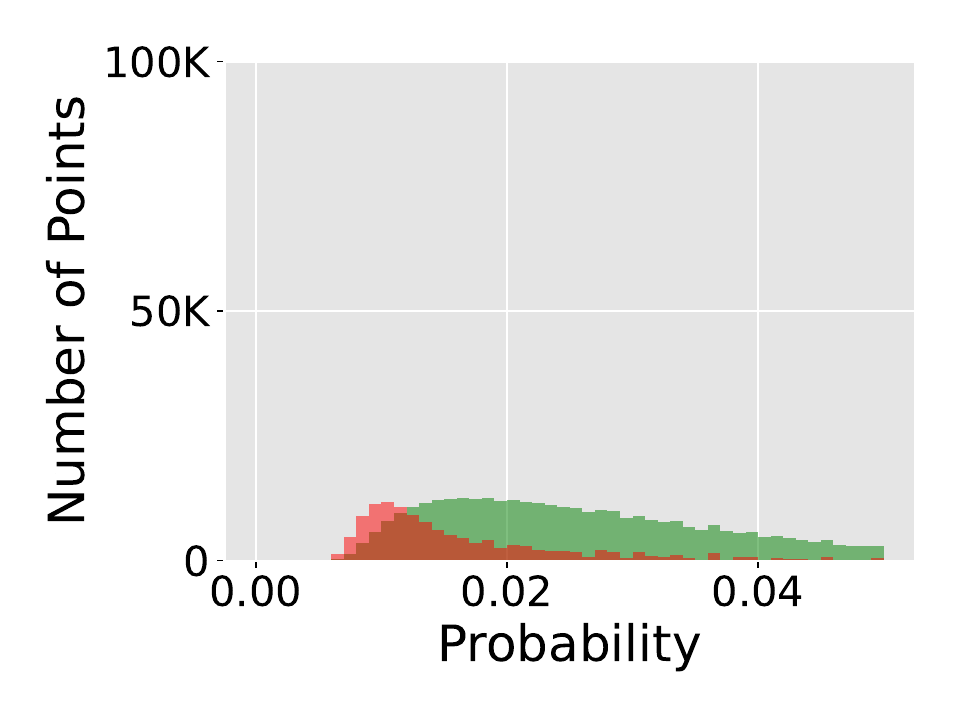}
\end{subfigure}
\hspace{-10pt}
\begin{subfigure}{0.25\linewidth}
\includegraphics[width=\linewidth]{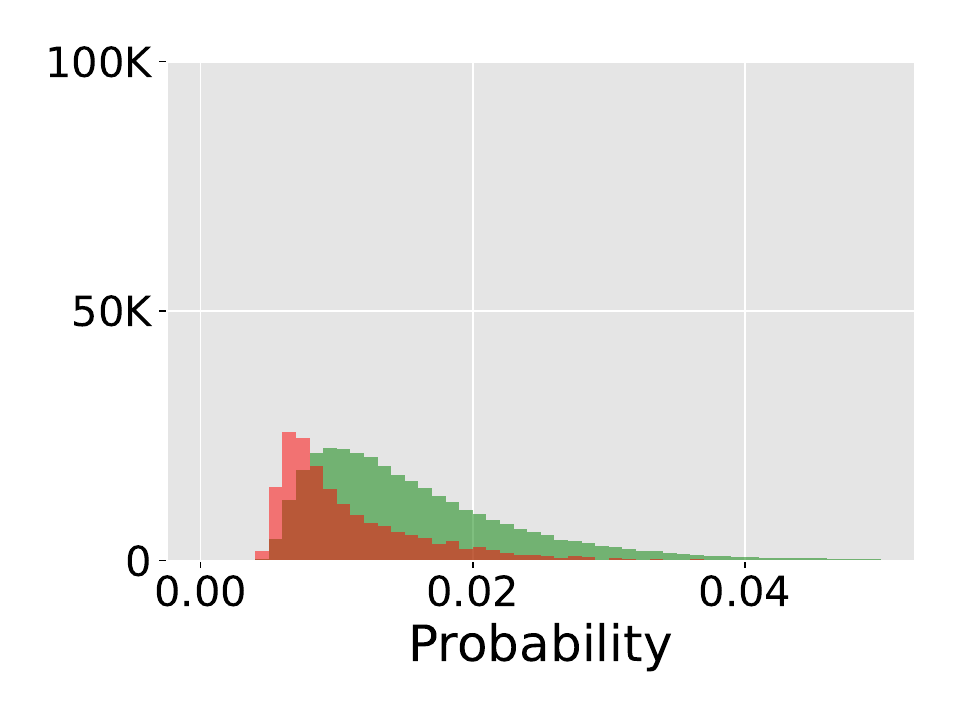}
\end{subfigure}
\hspace{-10pt}
\begin{subfigure}{0.25\linewidth}
\includegraphics[width=\linewidth]{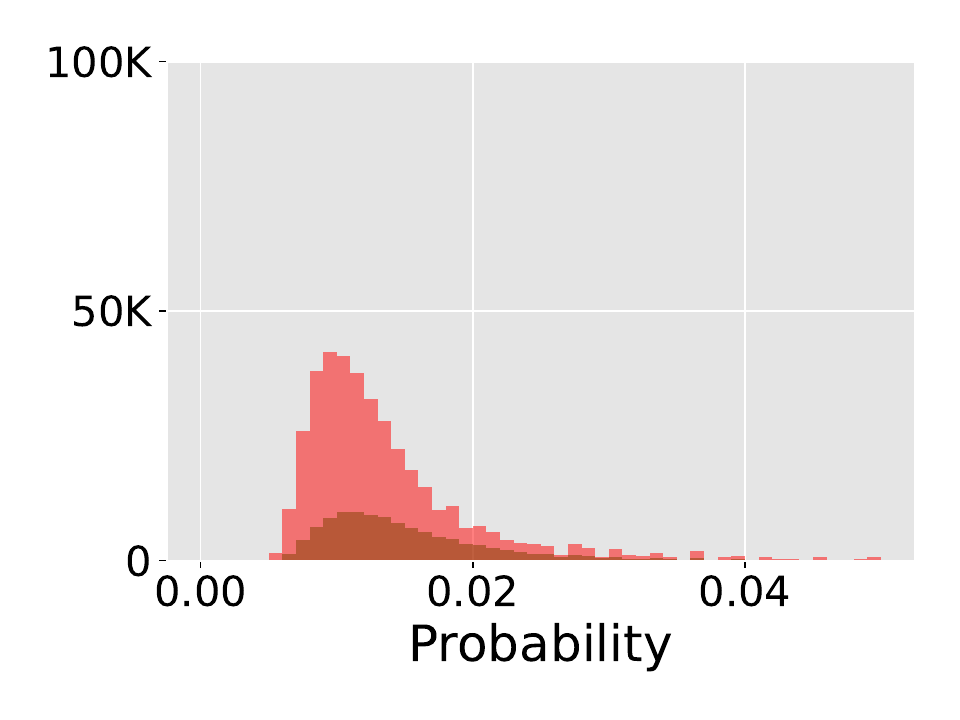}
\end{subfigure}
\hspace{-10pt}
\begin{subfigure}{0.25\linewidth}
\includegraphics[width=\linewidth]{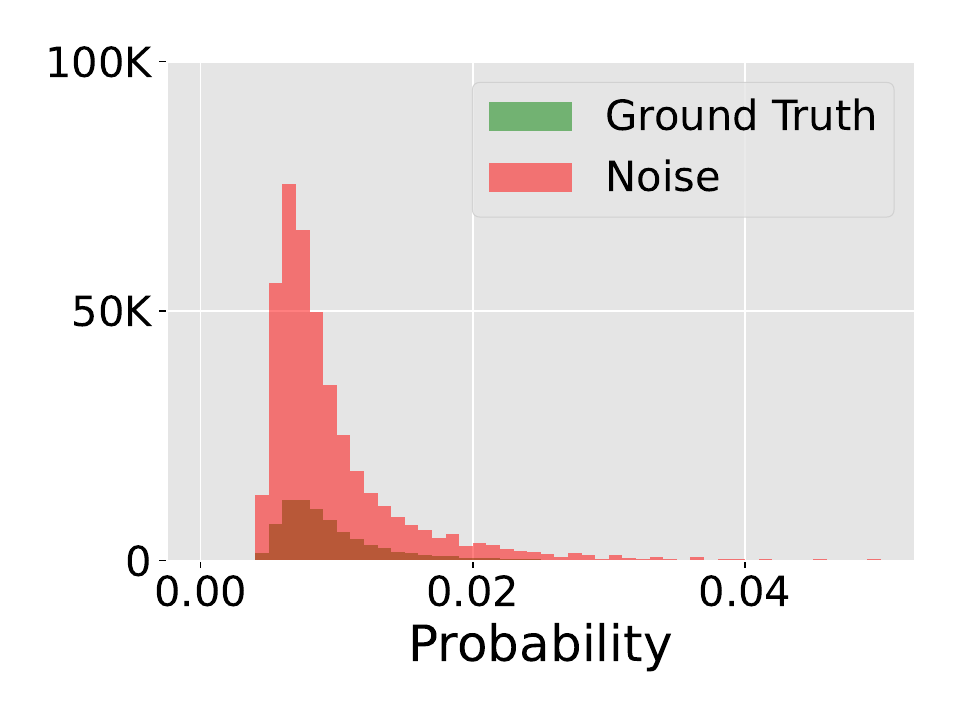}
\end{subfigure}

\vspace{-6pt}

\begin{subfigure}{0.25\linewidth}
\includegraphics[width=\linewidth]{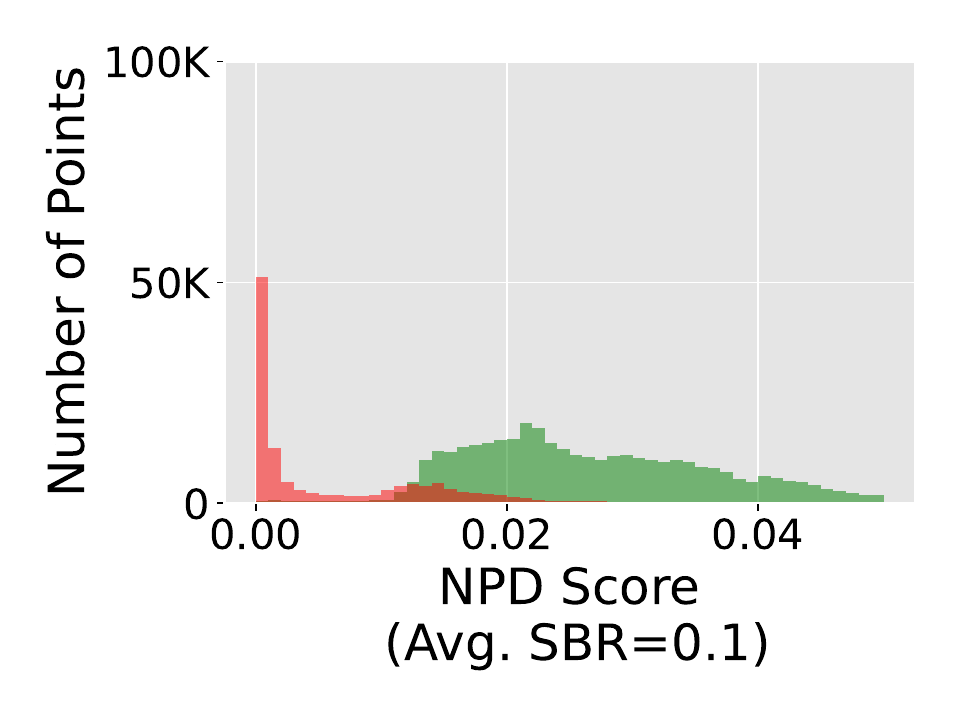}
\end{subfigure}
\hspace{-10pt}
\begin{subfigure}{0.25\linewidth}
\includegraphics[width=\linewidth]{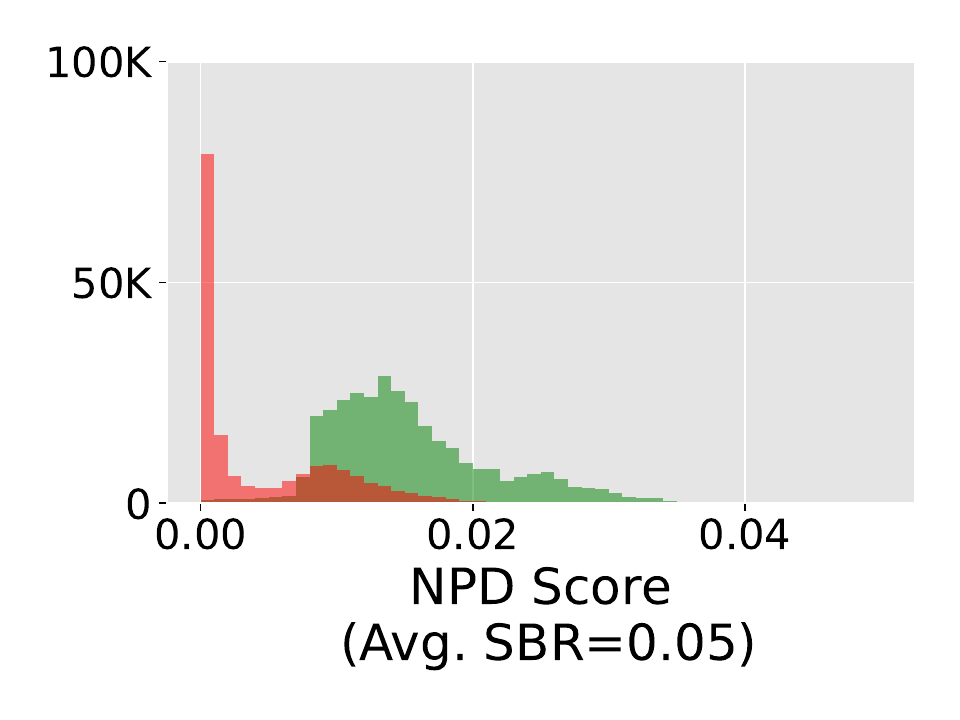}
\end{subfigure}
\hspace{-10pt}
\begin{subfigure}{0.25\linewidth}
\includegraphics[width=\linewidth]{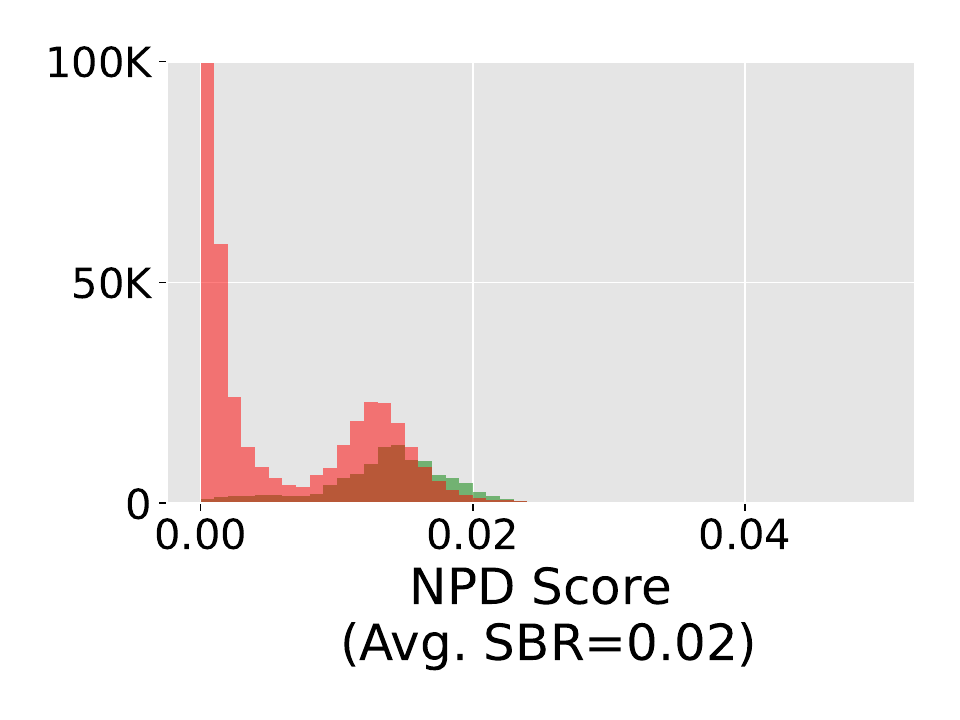}
\end{subfigure}
\hspace{-10pt}
\begin{subfigure}{0.25\linewidth}
\includegraphics[width=\linewidth]{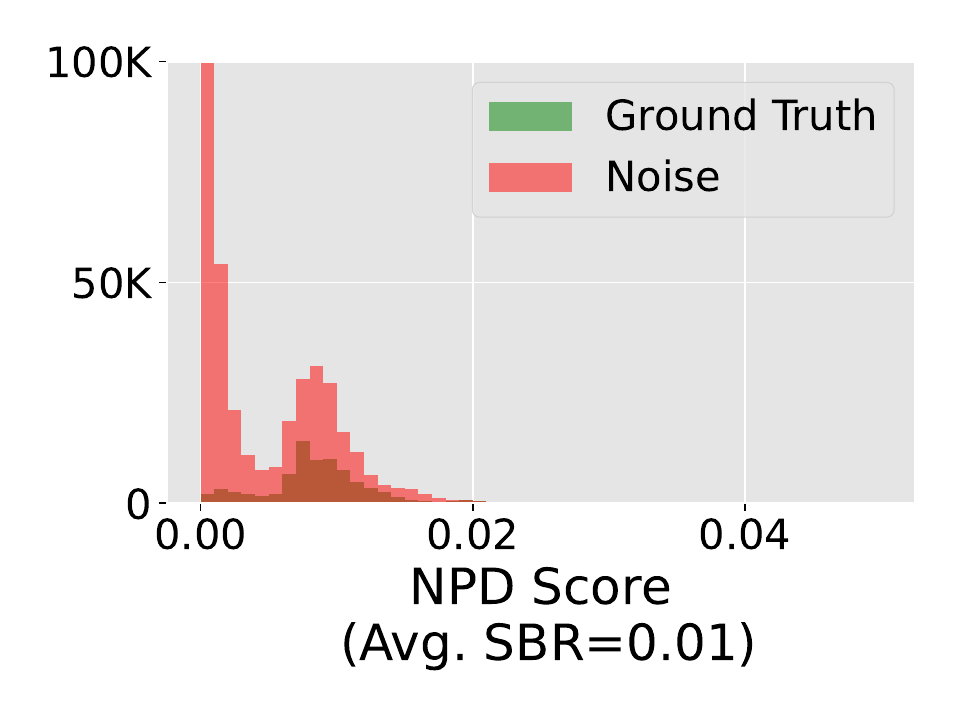}
\end{subfigure}
\vspace{-10pt}
\caption{\textbf{Distribution of Point Probability and NPD Score:} Point probabilities (top row) and NPD scores (bottom row) of all points in scenes under different SBR levels.
Point probability (proportional to the number of photon detections) alone is not a good indicator for separating noise from GT, as many GT points also have low probability.
NPD Scores for the noise (red) are lower as they have lower spatial density than ground truth points (green).
We use an NPD threshold to filter out many noise points from the point cloud.
}
\label{fig:npd_score_distribution}
\vspace{-7pt}
\end{figure*}


\medskip

\noindent \textbf{3D Sensing in Challenging Conditions}: 
Challenging sensing conditions discussed in Section 1 result in a timing histogram of low signal-to-background ratio (SBR).
Fig.~\ref{fig:noise_ppc} shows an example histogram of an ideal high SBR condition with a clear signal peak. Whereas in a challenging low SBR condition, there are multiple small peaks.
Consequently, the estimated depths will be noisy due to the lack of a dominant peak. Another challenge faced by SPADs in ambient illumination is that of photon pileup~\cite{heide2018sub,gupta2019asynchronous}, which distorts the histograms. There are approaches that computationally mitigate the structured pileup distortions~\cite{coates1968correction}, but end up amplifying the noise in the histograms~\cite{lee2023caspi}, making it challenging to detect low SBR peaks. In that regard, the proposed approaches are complementary to (and can be applied in conjunction with) these pileup mitigation methods.


\medskip

\noindent {\bf Noise in Low SBR Point Clouds:}
Noisy depth measurements under challenging conditions result in point clouds that are sparse or prone to severe noise. 
Since background peaks are uniformly distributed over the timing bins~\cite{rapp2017few}, the noise points are often arbitrarily far away from the ground truth (GT) depth and along the camera ray axis. 
Therefore, physically realistic noise under such scenarios is \emph{anisotropic} and often more severe than traditional isotropic Gaussian noise often considered in the existing literature.


\section{Probabilistic Point Clouds}
We introduce a confidence measure that can be derived from raw timing histograms using light-weight compute operations.
The following equation shows a probability $Pr(.)$ of a point defined as the ratio of photon detections for the peak bin to the total photon detections in a histogram:
\vspace{-5pt}\begin{equation}
\begin{split}
Pr(p^{ij}) & = \frac{h_{i,j}[m]}{ \sum_{n=1}^{N} h_{i,j}[n]},\\ 
\textnormal{where} \quad m & = \argmax_n\ h_{i,j}[n] 
\end{split}
\label{eq:probability_measure}
\end{equation}\vspace{-10pt}

\noindent and $p^{ij}$ is the point corresponding to the sensor pixel $(i,j)$ and $N$ is the number of timing bins in the histogram. We augment points with this probability attribute to create a Probabilistic Point Cloud (PPC).
Under ideal conditions and with no background photons, all photons would be detected in a single bin, resulting in a point with a probability of 1.
Fig.~\ref{fig:noise_ppc}c shows a PPC captured under a challenging scenario.
Our probability measure is a simple yet effective statistical estimate of the confidence in sensor depth measurements.

\subsection{Inference with Probabilistic Point Clouds}
The point-wise probability attribute in PPC provides vital information for robust inference.
Normally, one would expect most spurious points to have a low probability (due to smaller peaks from background photons). However, under challenging scenarios, even ground truth points may have low probability values. Hence, simply filtering low-probability points would remove noise, but also remove true scene points with low signal. To this end, we propose the following approaches to leverage the probability measure of points for robust inference.
A key feature of these approaches is that they do not require significant modifications to the inference model or its training procedures.

\smallskip

\subsubsection*{Neighbor Probability Density (NPD) Filtering}
Our key observation is that most spurious points in a point cloud under challenging conditions have low probability and/or low spatial density. In contrast, points that belong to true scene objects typically have a high local spatial density of points due to neighboring points being on the same surface or object.
To leverage this, we compute a score called the Neighbor Probability Density (NPD) score for each point, which encapsulates both the spatial density and the average probability of its neighbors as follows:
\vspace{-2pt}\begin{equation}
NPD(p_i) = \Sigma_{ p_j \in \mathcal{BQ}_{L,r}(p_i)} Pr(p_j) / L,
\label{eq:neighborscore}
\end{equation}
where $Pr(\cdot)$ is the probability of a point and $\mathcal{BQ}_{L,r}(\cdot)$ returns up to $L$ points that are in the local neighborhood ball of radius $r$ around a point. We aggregate the probability of these neighbors and normalize it with $L$ to get the final score. Since the ball query returns up to $L$ neighbors, the score for spatially dense points with $>L$ neighbors is the average probability of its $L$ neighbors in the local neighborhood ball, whereas sparse points with $<L$ neighbors get penalized with a lower score, as we normalize the score with $L$ which is greater than its number of neighbors.

Fig.~\ref{fig:npd_score_distribution} shows the distribution of NPD scores of all points in scenes under different SBR levels.
A large peak of noise points (red) on the left of each plot has much lower NPD scores than the GT points (green). The points below a certain NPD score ($\alpha$) can be filtered out without removing many GT points. As SBR decreases, GT points also have lower NPD scores, as expected, but the separation between the peak of noise points and the rest remains. Another smaller peak towards the right also contains some noise points with higher NPD scores. NPD filtering cannot remove these points, as it would also remove many GT points.

NPD score is a simple but effective way to filter out noise points as it leverages information from multiple sensor measurements by considering neighboring points, whereas filtering approaches based on low photon counts only rely on the timing histogram of the same pixel.
Note that we do not make any assumptions regarding the object/scene surface, and the score is easy to calculate without much computational overhead.

\begin{figure}[tp!]
\vstretch{0.95}{\includegraphics[width=\linewidth]{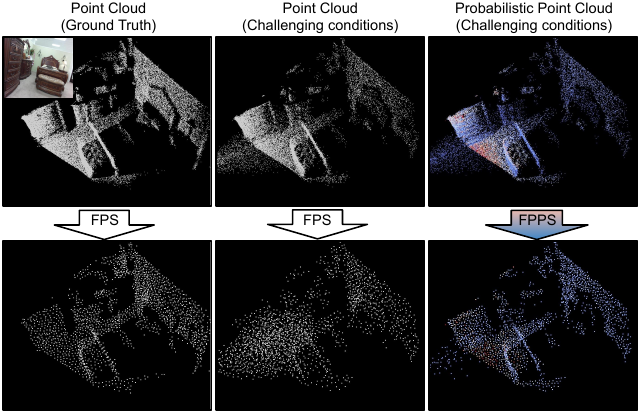}}
\vspace{-15pt}
\caption{\textbf{Farthest Probable Point Sampling (FPPS):} Farthest Point Sampling (FPS) is not robust to the noise under challenging low SBR conditions. It samples a large number of noise points as it prioritizes farther points (column 2). We propose FPPS, which only uses a candidate set of high-probability points for sampling. This ensures most sampled points are on the object surface while still covering the entire scene (column 3).}
\label{fig:fps_sampling}
\vspace{-8pt}
\end{figure}

\subsubsection*{Farthest Probable Point Sampling (FPPS)}
A key component in many point cloud inference models (\eg, PointNet++~\cite{qi2017pointnet++} and Point Transformer~\cite{zhao2021point}) is the Farthest Point Sampling (FPS), which is used for keypoint sampling. FPS samples the points that are farthest from each other. This ensures that sampled keypoints cover all scene regions, including objects with sparse or few points.


However, FPS is not robust to strong noise as it prioritizes points that are distant from each other. Since noise points can be spread out far from object surfaces, this results in a large number of noise points being sampled and thus a significant drop in performance.
Fig.~\ref{fig:fps_sampling} shows a point cloud of a scene under both ideal (column 1) and challenging (column 2) conditions, along with its sampled keypoints using FPS.
While FPS is effective in high SBR scenarios and covers most surfaces and objects, it suffers in challenging scenarios by sampling a lot of noise.

To address this issue, we propose Farthest \emph{Probable} Point Sampling (FPPS), which only considers high-probability points as candidates for sampling.
We build a candidate set of points above a certain probability value ($\beta$) and perform FPS on this candidate set.
This ensures that the sampled centers have fewer spurious points and more surface or ground truth points, which results in more effective inference output
(Fig.~\ref{fig:fps_sampling}).


We should note that the low-probability points are still part of the point cloud and are included in the rest of the network operations, like feature aggregation.
This allows the network backbone to still utilize low-probability points for feature extraction if they are in the neighborhood of a sampled keypoint.
FPPS is not needed for the network backbones that do not include a keypoint sampling operation.



\section{3D Object Detection Results}
\label{sec:3ddetection}

\noindent {\bf Datasets:}
We evaluate our approach on 3D object detection benchmarks of SUN RGB-D~\cite{song2015sun} and KITTI~\cite{geiger2012we}.
We use ground truth depth maps
to simulate a physically realistic photon timing histogram for each pixel using the simulation model and code provided by \cite{lindell2018single}.
Each histogram has 1024 bins and a temporal bin width of 97ps.
Our pulsed signal has a repetition period of 100ns, and the detected illumination pulse has a full-width half maximum of $\sim$350ps.
We simulate histograms with various levels of mean Signal to Background Ratio (SBR) for the scene to cover a variety of scenarios.
Our benchmark consists of the following SBR levels \{0.1 (5-50), 0.05 (5-100), 0.02 (1-50), 0.01 (1-100), \emph{Clean}\} where \emph{Clean} denotes the ground truth point cloud.

\medskip

\begin{table*}
\vspace{-5pt}
\footnotesize
\centering
\setlength\tabcolsep{6.7pt} 
\begin{tabular}{c ccc ccc ccc ccc cc}
\toprule
Avg. SBR & \multicolumn{2}{c}{Clean} && \multicolumn{2}{c}{0.1} && \multicolumn{2}{c}{0.05} && \multicolumn{2}{c}{ 0.02 } && \multicolumn{2}{c}{0.01} \\
\cline{2-3}\cline{5-6}\cline{8-9}\cline{11-12}\cline{14-15}
 & \scriptsize AP@25 & \scriptsize AP@50 && \scriptsize AP@25 &\scriptsize AP@50 &&\scriptsize  AP@25 &\scriptsize AP@50 &&\scriptsize AP@25 &\scriptsize AP@50 &&\scriptsize AP@25 &\scriptsize AP@50\\
\midrule
Matched Filtering & 51.34 & 27.45 && 42.43 & 20.49 && 38.77 & 17.57 && 16.95 & 5.05 && 11.34 & 2.73\\
Thresholding  & 57.11 & 33.21 && 51.27 & 28.62 && 46.44 & 24.86 && \underline{29.58} & \underline{14.81} && \underline{16.47} & \underline{6.45}\\
PointClean Net~\cite{rakotosaona2020pointcleannet} & 54.58 & 31.89 && 45.65& 26.44 && 40.19& 19.15 && 18.24 & 8.05 && 12.78 & 3.01\\
Score Denoising~\cite{luo2021score} & \underline{57.38} & \underline{34.02} && \underline{53.19} & \underline{29.45} && \underline{48.61} & \underline{25.78} && 26.35 & 13.73 && 14.55 & 4.73\\
PathNet~\cite{wei2024pathnet} & 57.16 & 33.87 && 52.16 & 28.79 && 47.11 & 24.89 && 25.45 & 12.96 && 13.87 & 4.56\\
\textbf{PPC (Ours)} & \textbf{58.61} & \textbf{34.99} && \textbf{54.29} & \textbf{31.15} && \textbf{52.46} & \textbf{30.20} && \textbf{38.49} & \textbf{16.47} && \textbf{29.42} & \textbf{13.16}  \\
\bottomrule
\end{tabular}
\vspace{-5pt}
\caption{\textbf{3D Object Detection Comparison}: Table shows AP@25 and AP@50 results on the SUN RGB-D dataset using VoteNet architecture.
Our approach outperforms all baselines and shows large gains under very low SBR conditions.}
\label{tab:3ddetectionresults}
\end{table*}

\begin{figure*}
\vspace{-5pt}
\center{
\vstretch{0.90}{\includegraphics[width=\linewidth]{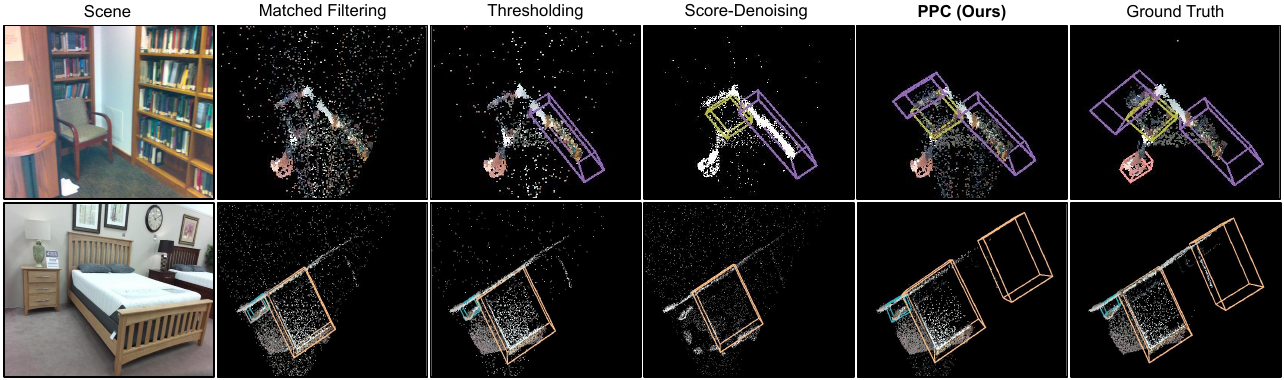}}}
\vspace{-15pt}
\caption{\textbf{3D Object Detection Visualization:} Figure shows results from the SUN RGB-D dataset using VoteNet under low SBR (0.05) conditions.
Matched Filtering struggles with noise and misses a lot of objects. Thresholding misses smaller (\textcolor{yellow}{chair}) and farther (\textcolor{blue-violet}{bookshelf}) objects. Score-Denoising removes noise closer to the surface but still misses a few objects.
PPC (Ours) outperforms all baselines.}
\vspace{-7pt}
\label{fig:results_5_100}
\end{figure*}

\noindent {\bf Implementation:}
We implement our method using the MMDetection3D framework~\cite{mmdet3d2020} by OpenMMLab.
For the SUN RGB-D dataset, we evaluate using the VoteNet \cite{qi2019deep} architecture, which is a LiDAR-based 3D object detector and uses PointNet++~\cite{qi2017pointnet++} as the point processing network backbone.
For the KITTI dataset, we use the PV-RCNN \cite{shi2020pv} architecture, which is a LiDAR-based 3D object detector and uses 3D Voxel CNN with sparse convolutions \cite{graham20183d} as a backbone.
Our NPD filtering step uses $L(=64)$ neighbors within radius $r(=0.2)$ to calculate the NPD score.
We use $\alpha$ = 0.003 and $\beta$ = 0.01 as hyperparameters for NPD Filtering and FPPS, respectively.
Please refer to the supplementary text for ablation studies on hyperparameters.

\medskip

\noindent {\bf Baselines:} We compare our approach with the following set of baselines. All methods use the same 3D detection model architecture and backbone for a fair comparison. 

\smallskip

\squishlist
\item \emph{Matched Filtering} \cite{1057571}: This method uses matched filtering output of the timing histograms. We convolve the histograms with the signal pulse in Eq.~\ref{eq:distance} before calculating the depth estimate. This provides a strong baseline for temporal denoising.

\item \emph{Thresholding:} This method uses a thresholding approach where depth estimates corresponding to small bin values are ignored. This removes a large number of spurious points from the point cloud. We select the optimal value of the threshold for our evaluation.

\item \emph{PointClean Net} \cite{rakotosaona2020pointcleannet}: This is a point cloud denoising network that uses a combination of outlier removal and denoising steps.

\item \emph{Score-based Denoising} \cite{luo2021score}: This is a state-of-the-art point cloud denoising approach, which denoises each point in the point cloud by updating it to its estimated local surface based on a calculated score.

\item \emph{PathNet} \cite{wei2024pathnet}: This is a point cloud denoising method based on reinforcement learning using a noise and geometry-based reward function.

\squishend

\smallskip

We use a matched filtering step for all methods (including PPC) for temporal denoising of the histograms.
All denoising networks (PointClean Net, PathNet, and Score-denoising) are retrained on SUN RGB-D dataset. Denoised point clouds are used for training and testing 3D detection models for these baselines.

\medskip

\noindent {\bf Benchmark Results:}
For every method, we train a joint model using all SBR levels and evaluate the performance at each SBR level.
Table~\ref{tab:3ddetectionresults} shows AP@25 and AP@50 comparisons on the SUN RGB-D dataset with VoteNet.
Matched Filtering and Thresholding suffer due to a large amount of noise.
PointClean Net, PathNet, and Score-denosing show improvement for higher SBR levels, but are not as effective under low SBR conditions.
Our approach performs significantly better even under extremely low SBR conditions.
Please see the supplementary text for per-category results.
Table~\ref{tab:3ddetectionresults_kitti} shows mAP on moderate difficulty of KITTI val split calculated with 11 recall positions for PV-RCNN architecture~\cite{shi2020pv}. Our approach shows significant gains for the pedestrian and cyclist categories under low SBR conditions.


\begin{table*}
\vspace{-5pt}
\centering
\footnotesize
\setlength\tabcolsep{4.1pt} 
\begin{tabular}{c ccc c ccc c ccc c ccc c ccc}
\toprule
Avg. SBR & \multicolumn{3}{c}{Clean} && \multicolumn{3}{c}{0.05} && \multicolumn{3}{c}{0.02} && \multicolumn{3}{c}{0.01} && \multicolumn{3}{c}{0.005} \\
\cline{2-4}\cline{6-8}\cline{10-12}\cline{14-16}\cline{18-20}
 & Car & Ped & Cyc &  & Car & Ped & Cyc & & Car & Ped & Cyc & & Car & Ped & Cyc & & Car & Ped & Cyc  \\
\midrule
Matched Filtering & 82.48 & 60.11 & 71.36 && \textbf{73.14} & 55.76 & 61.84 && 68.17 & 50.03 & 52.85 && 59.95 & 47.06 & 43.74 && 50.68 & 37.01 & 35.01 \\
Thresholding & 82.81 & 58.63 & 71.55 && 72.80 & 57.72 & 60.44 && 68.05 & 54.80 & 52.71 && 59.40 & 49.23 & 44.96 && 50.35 & 38.62 & 35.74\\
\textbf{PPC (Ours)} & \textbf{83.56} & \textbf{60.62} & \textbf{73.35} && 73.03 & \textbf{59.12} & \textbf{64.14} && \textbf{68.42} & \textbf{59.04} & \textbf{53.18} && \textbf{60.29} & \textbf{55.39} & \textbf{47.76} && \textbf{51.30} & \textbf{49.51} & \textbf{36.44}\\
\bottomrule
\end{tabular}
\vspace{-5pt}
\caption{\textbf{KITTI 3D Detection Comparison}: Table shows mAP for car, pedestrian, and cyclist categories on moderate difficulty of KITTI val split calculated with 11 recall positions for PV-RCNN architecture. Our method shows significant gains under low SBR conditions.}
\vspace{-10pt}
\label{tab:3ddetectionresults_kitti}
\end{table*}


\medskip

\begin{figure}[tp!]
\vspace{-2pt}
\includegraphics[width=\linewidth]{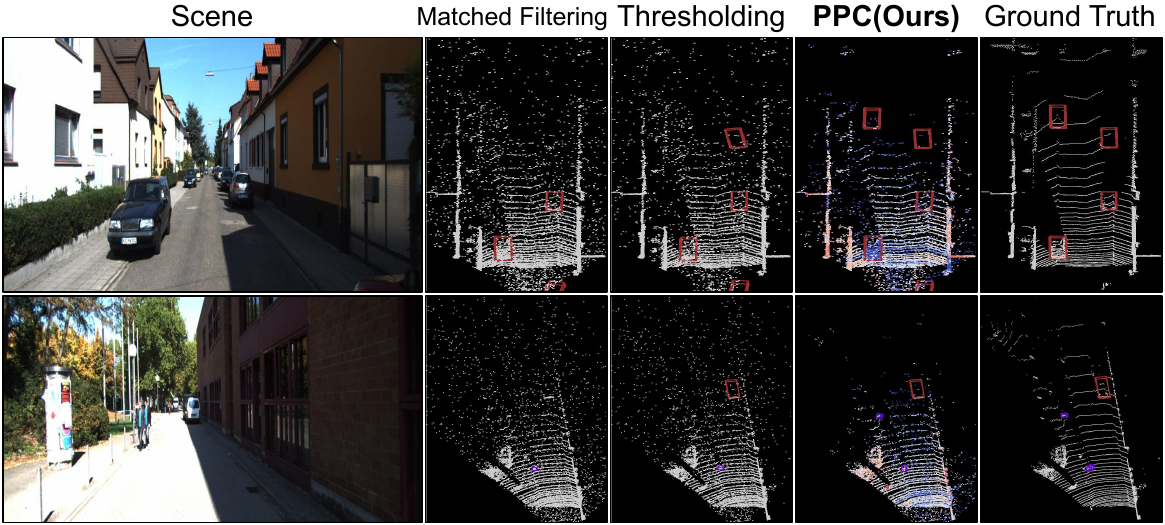}
\vspace{-16pt}
\caption{\textbf{3D object detection visualization} on the KITTI dataset under low SBR (0.02) conditions. Baselines miss small and farther objects: \textcolor{red}{car} and \textcolor{blue-violet}{pedestrian} (zoom-in). PPC detects most objects.}
\label{fig:kitti_results_1_50}
\end{figure}

\noindent \textbf{Observations:} 
Fig.\ \ref{fig:results_5_100} and \ref{fig:kitti_results_1_50} visualize results on the SUN RGB-D and KITTI datasets, respectively, using LiDAR-only 3D detectors.
Matched filtering misses many objects due to strong noise.
Thresholding removes a large amount of noise but also removes ground truth points that have low photon detections.
Thus, it performs better on larger and closer objects, as they include points with high incident photon flux, but misses farther objects like nightstands and chairs.
Score-based Denoising is able to denoise points closer to a surface.
However, the recognition is still affected by noise points that are far from the surface, as they are not denoised effectively.
Our approach can detect smaller and farther objects more consistently.
Color information is only used for visualization and not for inference.
We refer to the supplement for more visualization results and some failure modes, such as extremely low SBR histograms.

\medskip

\noindent \textbf{Generalization}:  We also evaluate using ImVoteNet~\cite{qi2020imvotenet} (fusion of camera and LiDAR), Uni3DETR~\cite{NEURIPS2023_7d60bfd8} (LiDAR only with Transformer-based architecture), and PointPillars~\cite{lang2019pointpillars} (LiDAR only with pillar-based representation) in the Table 4-6 (supplementary) which shows the effectiveness of our method on a wide range of 3D detectors.
Our approach can similarly be extended to other 3D inference tasks, like point cloud classification and segmentation.


\medskip

\noindent \textbf{Comparison with Point Cloud Denoising Networks:}
Most point cloud denoising networks are trained for surface reconstruction tasks and are not designed for downstream inference. Table \ref{tab:3ddetectionresults} shows that denoising point clouds does not significantly improve the 3D object detection performance for challenging conditions.
This is partly because such methods consider local isotropic Gaussian noise for training point cloud denoising models. 
However, the noise in low SBR scenarios also consists of spurious outlier points that are uniformly distributed, resulting in large depth errors~\cite{rapp2017few}.
Further, denoising networks add a significant computational overhead.
Table~\ref{tab:inference_time} shows per-scene runtime for each method.
This suggests that training a robust inference model to handle noise is more beneficial than denoising point clouds under challenging conditions. 

\begin{table}[tp!]
\vspace{-2pt}
\centering
\setlength\tabcolsep{3.2pt} 
\scriptsize
\begin{tabular}{c c c c c c c}
\toprule
& Matched & Thresh- & PointClean & Score- & PathNet & PPC \\
& Filtering & olding & Net & Denoising & & (Ours)\\
\midrule
Runtime (ms) & 87 & 89 & 755 & 1345 & 867 & 95 \\
\bottomrule
\end{tabular}
\vspace{-5pt}
\caption{\textbf{Comparison of Inference Time:} Our method adds no significant computational cost while outperforming all baselines.}
\vspace{-7pt}
\label{tab:inference_time}
\end{table}

\medskip

\noindent \textbf{Comparison with Histogram Denoising Methods}: Our approach is complementary to the histogram denoising methods used for 3D reconstruction and can be used alongside such methods.
We evaluate PPC using denoised histograms from 3D-CNN \cite{peng2020photon} in Table 7 of the supplementary text, which shows $\sim$3\% mAP gain.
Please refer to the supplementary text for more details.

These reconstruction techniques require full readout of timing histograms, which incurs prohibitive bandwidth costs ($\sim$10s of GB/s), even for moderate resolution LiDARs \cite{gutierrez2022compressive}. While full histograms do offer richer information and higher performance, they are less practical for real-time, latency-critical applications such as 3D object detection. Our approach estimates a compact, simple representation derived from full histograms that uses several orders of magnitude smaller data rates ($\sim$10s of MB/s).

\subsection{Modeling with Probabilistic Points}

Moving forward, we explore the integration of the PPC representation directly into the point cloud models. To this end, we focus on VoteNet~\cite{qi2019deep} --- a classic model built on PointNet++~\cite{qi2017pointnet++} for 3D object detection. VoteNet~\cite{qi2019deep} processes a point cloud by taking points and their attributes as input, leveraging PointNet++~\cite{qi2017pointnet++} to extract point-wise features. These features are passed through a voting module that groups the point cloud into local clusters. Each cluster generates 3D boxes as the object proposal. The proposals are then classified to produce the final detection results.

This design allows integrating point probability at various stages of the model, including the input, point-wise features, and object proposals. We explore these options through the experiments detailed below. While alternative approaches to integrate probability into point cloud networks may exist, our goal is to delineate evident design choices to build PPC-aware 3D inference models.

\squishlist
\item \emph{Probability as a point attribute (A):} Point probability can be treated as an attribute for the point and used as an input for the network. This design tasks the network to learn effective features from the probability.
\item \emph{Probability weighted point feature vectors (B):} Point-wise features from PointNet++ can be weighted by the average neighborhood probability. This allows the network to prioritize features from high-probability points.
\item \emph{Probability weighted objectness scores (C):}
The objectness score for a proposal can be weighted by the average probability of points within the corresponding 3D box. This allows the network to assign higher objectness scores for proposals with high confidence points.
\squishend

\begin{figure*}[thp!]
\centering
\vspace{-5pt}
\vstretch{0.85}{\includegraphics[width=\linewidth]{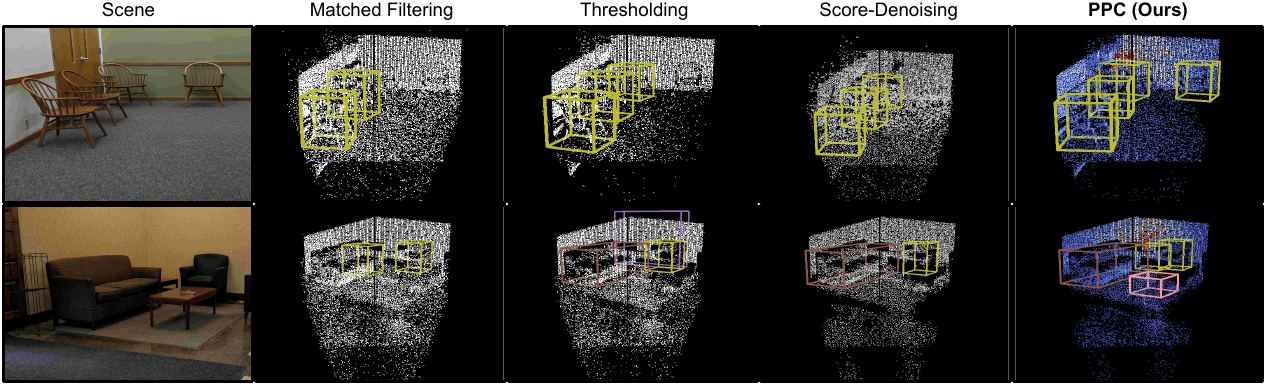}}
\vspace{-17pt}
\caption{\textbf{3D Detection Results on Indoor Real Captures:} Figure shows scenes under challenging conditions captured by our FLIMera LiDAR system.
Baselines fail to detect many small distant objects. PPC detects all objects (\eg \textcolor{Goldenrod}{chair}, \textcolor{Lavender}{table} and \textcolor{RawSienna}{couch}).}
\label{fig:flimera_results}
\vspace{-7pt}
\end{figure*}

\begin{figure*}[thp!]
\centering
\vstretch{0.85}{\includegraphics[width=\linewidth]{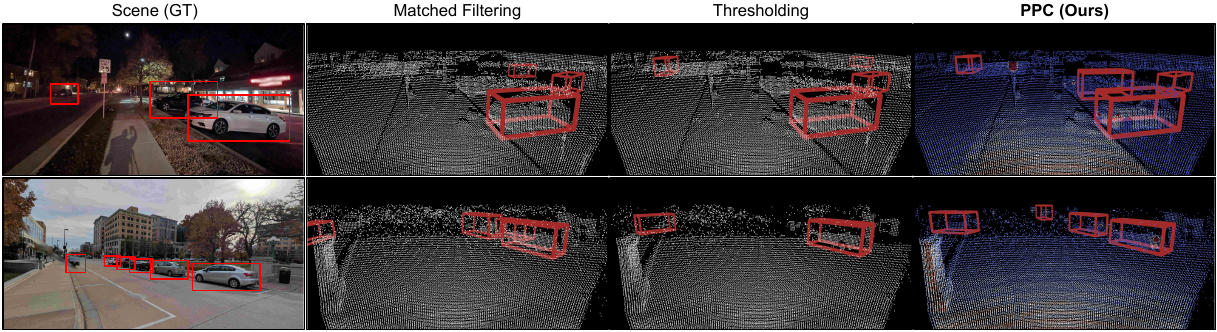}}
\vspace{-17pt}
\caption{\textbf{3D Detection Results on Outdoor Real Captures:} The Figure shows scenes under challenging conditions captured by our Adaps LiDAR system. Baselines fail to detect distant cars, whereas PPC detects farther \textcolor{red}{cars} with accurate bounding boxes.}
\label{fig:adaps_results}
\vspace{-9pt}
\end{figure*}

\medskip

\noindent \textbf{Results and Discussion:} Table~\ref{tab:3ddetectionresults_sunrgbdextra} presents the AP@25 results on the SUN RGB-D dataset. By employing individual options, the model attains a further gain of 1-2\%. Using all three options leads to a modest boost of $\sim$2\%. We note that this boost is on top of our already strong baseline. Our results show the potential of integrating PPC with deep models and further call for innovative approaches in this area.   



\begin{table}[tp!]
\centering
\footnotesize
\begin{tabular}{l c c c c c}
\toprule
Avg. SBR & Clean & 0.1 & 0.05 & 0.02 & 0.01 \\
\midrule
PPC (Baseline) & 58.61 & 54.29 & 52.46 & 38.49 & 29.42 \\
PPC + A & 58.35 & 55.73 & \underline{53.67} & \underline{39.10} & \underline{30.58} \\
PPC + B & \underline{59.29} & \underline{55.86} & 52.95 & 37.68 & 30.42\\
PPC + C & 58.53 & 55.18 & 53.15 & 38.76 & 30.17 \\
PPC + A+B+C & \textbf{59.35} & \textbf{56.11} & \textbf{53.45} & \textbf{39.87} & \textbf{30.81}  \\
\bottomrule
\end{tabular}
\vspace{-5pt}
\caption{AP@25 results on SUN RGB-D with PPC variants.}
\vspace{-10pt}
\label{tab:3ddetectionresults_sunrgbdextra}
\end{table}

\section{Experiments with Real Hardware}
Finally, we demonstrate our approach using real PPCs captured by our prototype Single-photon LiDAR systems.

\medskip

\begin{figure}[tp!]
\vspace{-2pt}
\centering
\vstretch{1.2}{\includegraphics[width=\linewidth]{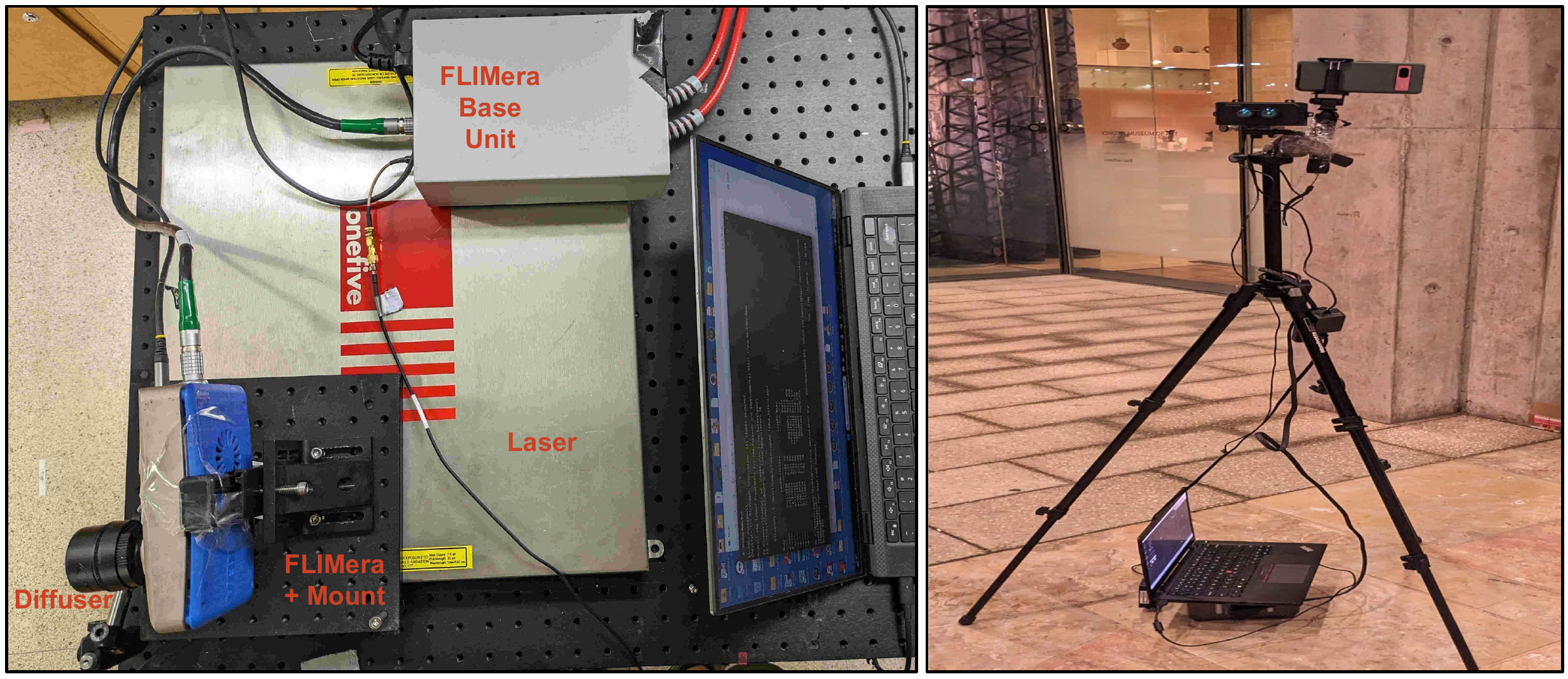}}
\vspace{-16pt}
\caption{\textbf{Camera Setup:} Figure shows our complete LiDAR setups with FLIMera (left) and Adaps (right) sensors.}
\label{fig:flimera_setup}
\vspace{-10pt}
\end{figure}

\noindent {\bf Indoor Capture Setup:} We use a HORIBA FLIMera~\cite{8681087} SPAD camera (Fig.~\ref{fig:flimera_setup}) as it allows access to raw timing histograms for each pixel. The camera consists of a 192x128 pixel SPAD array; each pixel has a quantized 12-bit (4096 bins) time axis, with each bin having a width of 41.1ps.
We consider several indoor scenes (\eg, conference rooms, lecture halls, and living rooms) under ambient light ranging from 200-800 lux under varying exposure times from 0.1s to 1s to simulate various signal levels.

\medskip

\noindent {\bf Outdoor Capture Setup:} We use Adaps ADS6311~\cite{adapsphotonics} sensor which is a commercial medium-range Single-photon LiDAR.
It has a spatial resolution of 256x192 pixels, where each pixel has 672 temporal bins with a bin width of 297ps.
We consider outdoor scenes (\eg, parking lots, traffic stops, and busy roads) for captures.
Please refer to the supplementary text for more details on both setups.

\medskip

\noindent {\bf 3D Object Detection Results:}
Figures \ref{fig:flimera_results} and \ref{fig:adaps_results} show the results of our approach compared with the baselines using real captures for indoor and outdoor scenes, respectively.
PPC can detect most objects with accurate bounding boxes. More results on real captures are in the supplementary text. 

\vspace{-2pt}
\section{Discussion and Future Outlook}
\vspace{-2pt}

In this work, we propose retaining the uncertainty in depth measurements from LiDAR-based 3D sensors.
This enables augmenting point clouds with a valuable probability attribute, which is easily computable on a sensor chip.
We explore methods for integrating this probabilistic point cloud into a 3D inference pipeline, such as object detection. Our approach is more robust on distant low-albedo objects, which we demonstrate using both existing datasets and real-world LiDAR captures.

\smallskip

\noindent \textbf{Confidence Measure from Other Depth Sensors:} Although this work only considers LiDARs, other 3D sensors such as stereo \cite{zaarane2020distance}, structured light \cite{gupta2013structured}, and indirect time-of-flight, \eg Azure Kinect \cite{qiu2019deep}, also suffer from noise and low fidelity in challenging imaging scenarios, and could benefit from similar probability attributes. Since the proposed inference approaches do not make any assumption about the nature of the probability attributes or the noise characteristics of the underlying sensing modality, a promising future research direction is to extend this work to a wide range of 3D sensors.

\smallskip

\noindent \textbf{Richer Confidence Measure:} 
We designed a simple probability attribute based on histogram peak locations, aiming for low computational and memory cost to support on-chip implementation. Future works could incorporate richer information of uncertainty, such as Cramér–Rao uncertainty estimates~\cite{altmann2015spectral}. Another interesting direction is to learn task-aware confidence measures for downstream inference, which could enhance the performance and generalizability across different LiDAR models.



\noindent \textbf{Acknowledgments:}
This research was supported by the National Science Foundation via CAREER Award \#1943149, \#2107060 (CNS), \#2333491 (CPS Frontier), \#2442739 (CAREER Award), by the Office of Naval Research via grant N000142412155, and by the Army Research Lab under contract number W911NF-2020221.
The authors also thank Atul Ingle for his insights.

{
    \small
    \bibliographystyle{ieeenat_fullname}
    \bibliography{main}
}

\clearpage
\setcounter{page}{1}
\setcounter{figure}{0}
\setcounter{table}{0}
\setcounter{section}{0}
\maketitlesupplementary

We provide additional details and results that complement the main paper in this report. We include more results with real captures (section 1), ablation studies (section 2), implementation details and additional results from our 3D detection benchmark (section 3), results with more 3D detectors (section 4), histogram denoising and compression methods (section 5) and raw histogram examples under low SBR (section 6).

\section{Recognition with Real Captures}
In this section, we provide further details about our LiDAR setups and more results on real PPC captures.

\subsection{Camera Setups}
\begin{figure}[tp!]
\centering
\begin{subfigure}[justification=centering]{0.7\linewidth}
\frame{
\vstretch{0.9}{\includegraphics[width=\linewidth]{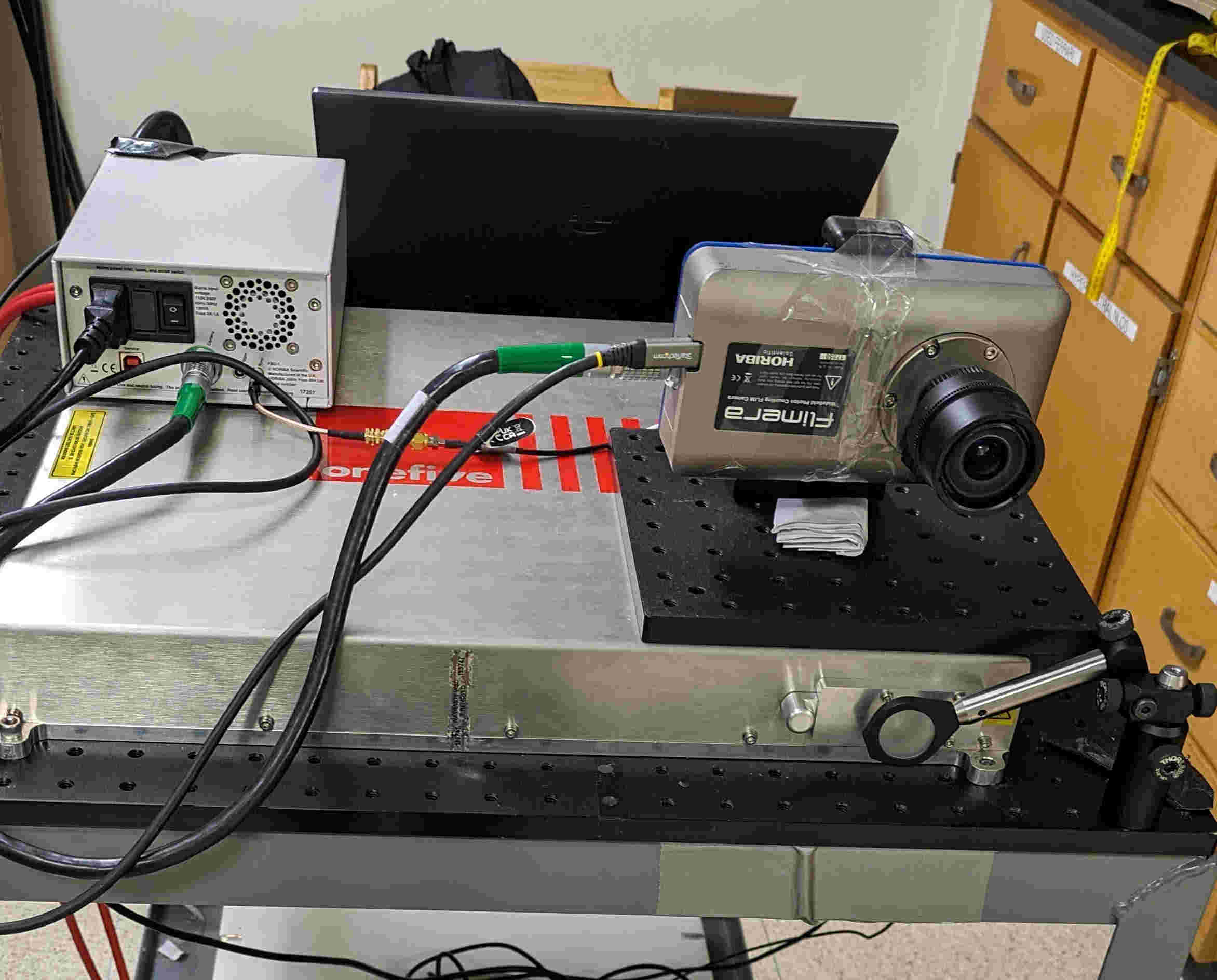}}}
\end{subfigure}
\begin{subfigure}[justification=centering]{0.2\linewidth}
\includegraphics[width=\linewidth]{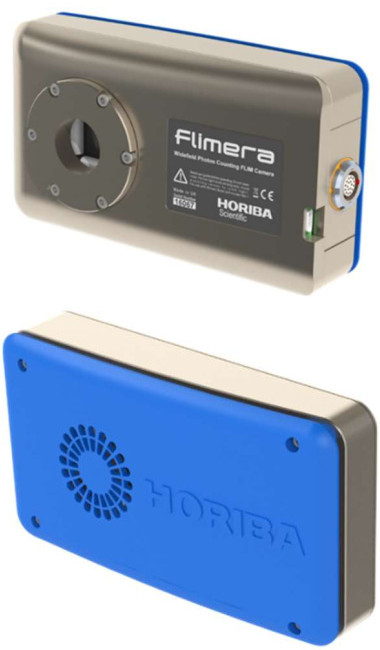}
\end{subfigure}
\vspace{-2pt}
\caption{
\textbf{Indoor Camera Setup:} The Figure shows the front view of our FLIMera camera setup (left) and the sensor (right).}
\label{fig:flimera_setup_side}
\vspace{-2pt}
\end{figure}

We use a LiDAR sensor with an external laser for our indoor captures.
This allows us to control various camera and scene parameters (\eg, exposure time, laser power, and ambient illumination) over a wide range.
We use a commercial LiDAR sensor for our outdoor captures.
This allows us to have a portable low-power LiDAR setup for outdoor environments. Here are the details of both setups:

\medskip
\noindent {\textbf{Indoor Camera Setup:}} Our indoor setup uses a SPAD-LiDAR sensor with an external class 4 laser.
Fig.~\ref{fig:flimera_setup_side} shows the front view of our setup with the HORIBA FLIMera~\cite{8681087} camera.
The temporal resolution of the camera is about 380ps, which is in line with the full-width at half-maximum (FWHM) of the instrument response function (IRF) of the device. 
We set up our camera system with the Katana laser~\cite{katanalaser}, which is a high-powered pulsed picosecond laser system by OneFive.
The laser has a wavelength of 532nm (green).
We operate the laser under low power settings (ranging from 50-100mW).
We use a flash illumination setup with a diffuser to illuminate the field of view of the sensor. 
We use a 3.8mm focal length lens for a wider field of view of the scene.
The laser system, as well as the FLIMera sensor, is connected to an external computer to receive the synchronization signal and trigger for the capture. 

\begin{figure}[tp!]
\centering
\begin{subfigure}[justification=centering]{0.5\linewidth}
\frame{\vstretch{0.7}{\includegraphics[width=\linewidth]{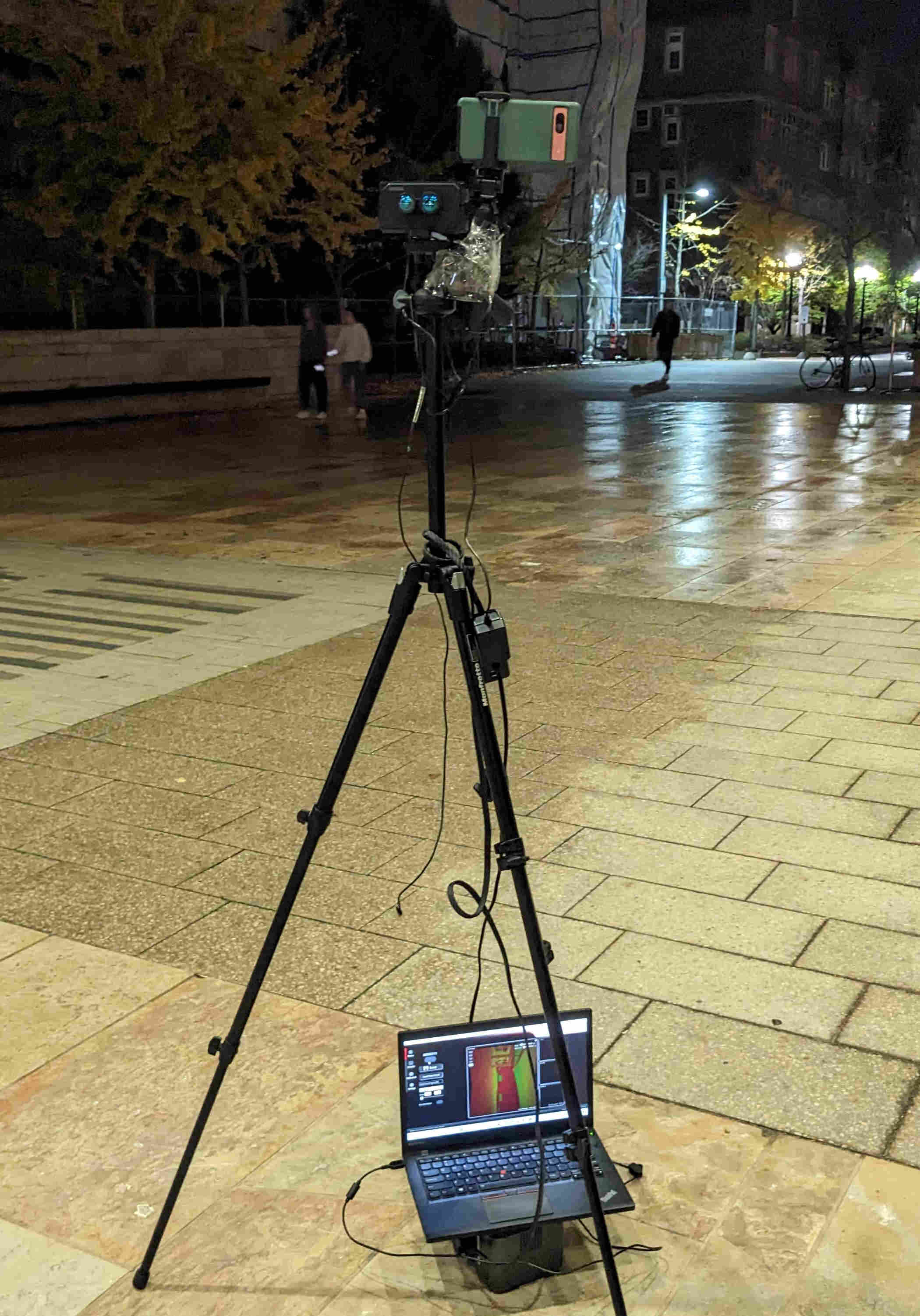}}}
\end{subfigure}
\begin{subfigure}[justification=centering]{0.25\linewidth}
\includegraphics[width=\linewidth]{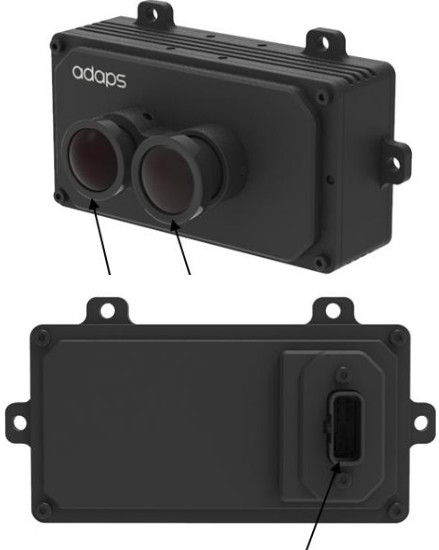}
\end{subfigure}
\vspace{-2pt}
\caption{
\textbf{Outdoor Camera Setup:} The Figure shows the front view of our Adaps camera setup (left) and the sensor (right).}
\label{fig:adaps_setup}
\vspace{-2pt}
\end{figure}

\medskip
\noindent {\textbf{Outdoor Camera Setup:}}
Our outdoor camera setup uses a commercial LiDAR sensor, which is a more portable camera and uses low input power. 
Fig.~\ref{fig:adaps_setup} shows the front view of the setup with Adaps~\cite{adapsphotonics} camera.
The camera is rated for an accuracy of more than 5cm up to a range of 30m. 
The setup has a wide FOV (120\textdegree horizontal and 90\textdegree vertical).
We also vary the exposure time from 0.1s to 1s to simulate various signal levels.
The camera can operate at a very low power input ($<$10W) and is connected to a small portable AC power source.
The camera also saves low-resolution (256x192) grayscale images, which are used for visualizations only.
We also mount a smartphone camera in our setup to simultaneously capture high-res RGB images, used for visualizations only for some static scenes shown in the main paper and supplementary report.

\subsection{3D Object Detection Results}

\begin{figure*}[htp!]
\centering
\vspace{-5pt}
\includegraphics[width=\linewidth]{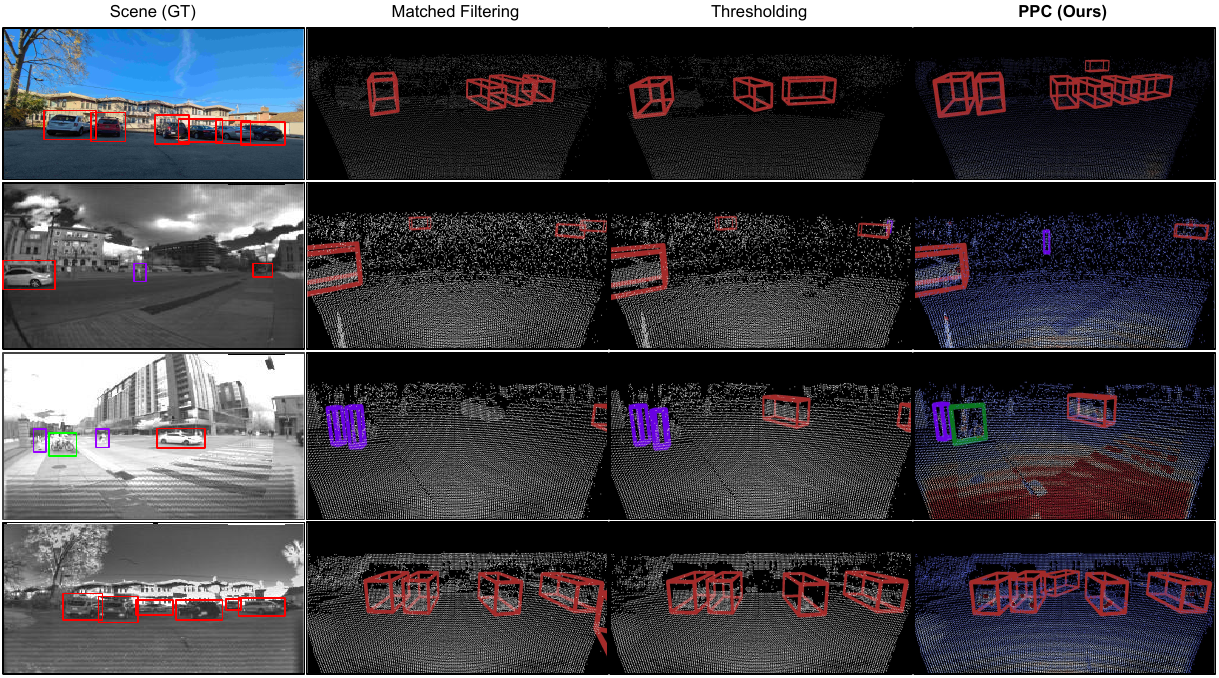}
\vspace{-15pt}
\caption{\textbf{3D Detection Results on Real Outdoor Captures:} Figure compares our method with the baselines under very challenging low SBR conditions. The first scene includes 6 cars, and baselines fail to detect farther cars. The second scene shows 2 cars and a pedestrian, and the baselines struggle to detect the distant pedestrian. The third scene includes a car, two pedestrians, and a cyclist. The ground truth objects are marked in the camera scene images for easier visualization. PPC detects most objects ($\eg$ \textcolor{red}{cars}, \textcolor{blue-violet}{pedestrians}, and \textcolor{green}{cyclists}) in all scenes with accurate bounding boxes.
}
\label{fig:adaps_results_supp}
\end{figure*}

\begin{figure*}[htp!]
\centering
\vstretch{0.90}{\includegraphics[width=\linewidth]{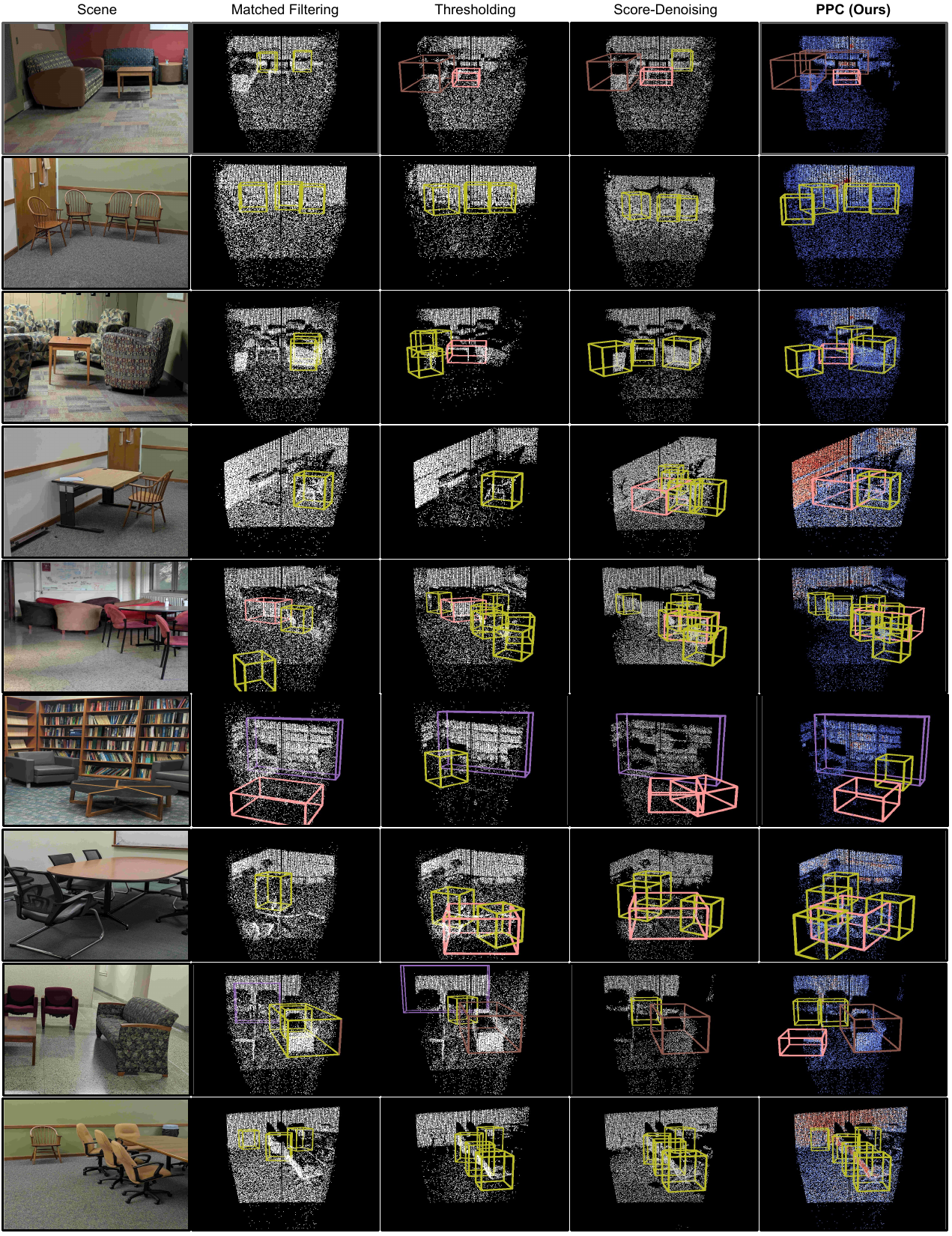}}
\vspace{-17pt}
\caption{\textbf{3D Detection Results on Real Indoor Captures:} Figure compares our method with the baselines under challenging low SBR conditions. Baselines fail to detect many small and farther objects, $\eg$ chairs in the back, in many scenes. PPC detects most objects (\eg \textcolor{Goldenrod}{chair}, \textcolor{Lavender}{table}, and \textcolor{RawSienna}{couch}) with tight bounding boxes.
}
\label{fig:flimera_results_supp}
\end{figure*}

Fig. \ref{fig:adaps_results_supp} and \ref{fig:flimera_results_supp} show a comparison of our approach with the baselines using real indoor and outdoor captures. 
Matched Filtering baseline suffers from noise and often detects false positives. Thresholding frequently misses small or farther objects in the scene. Baselines struggle with farther chairs in indoor captures and farther cars and pedestrians in outdoor captures.
Our approach detects most objects with tight bounding boxes.

\clearpage
\clearpage

\section{Ablation Studies}
In this section, we include the ablation studies of our proposed method. We evaluate various design choices on the SUN RGB-D benchmark using the VoteNet architecture.

\begin{figure}[htp!]
\centering
\vspace{-5pt}
\begin{subfigure}{0.49\linewidth}
\includegraphics[width=\linewidth]{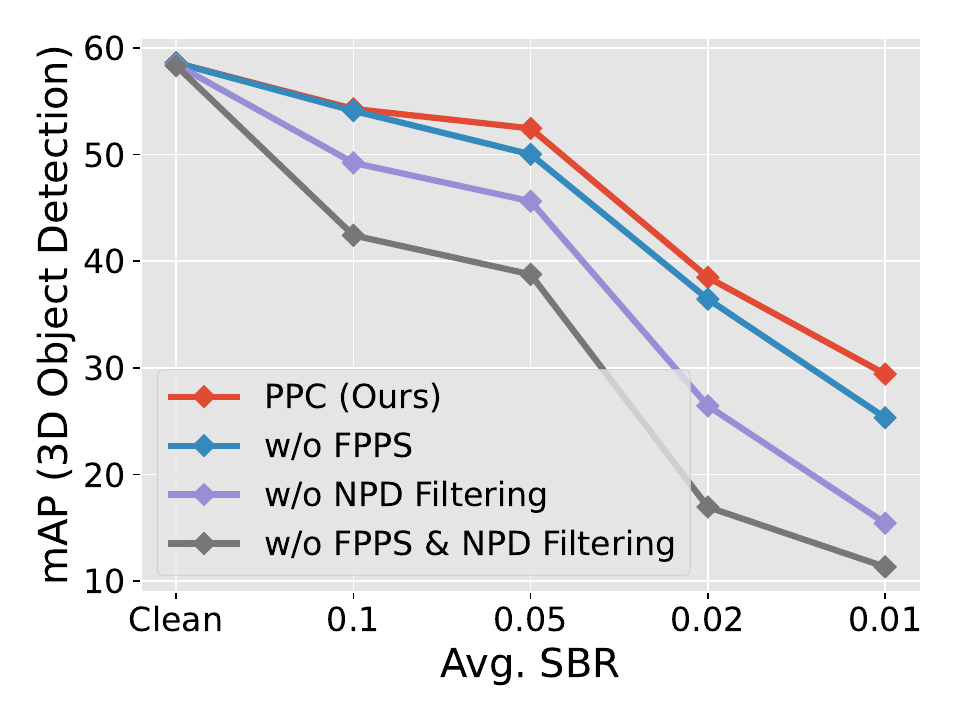}
\vspace{-15pt}
\caption{}
\label{fig:ablation_npdfps}
\end{subfigure}
\begin{subfigure}{0.49\linewidth}
\includegraphics[width=\linewidth]{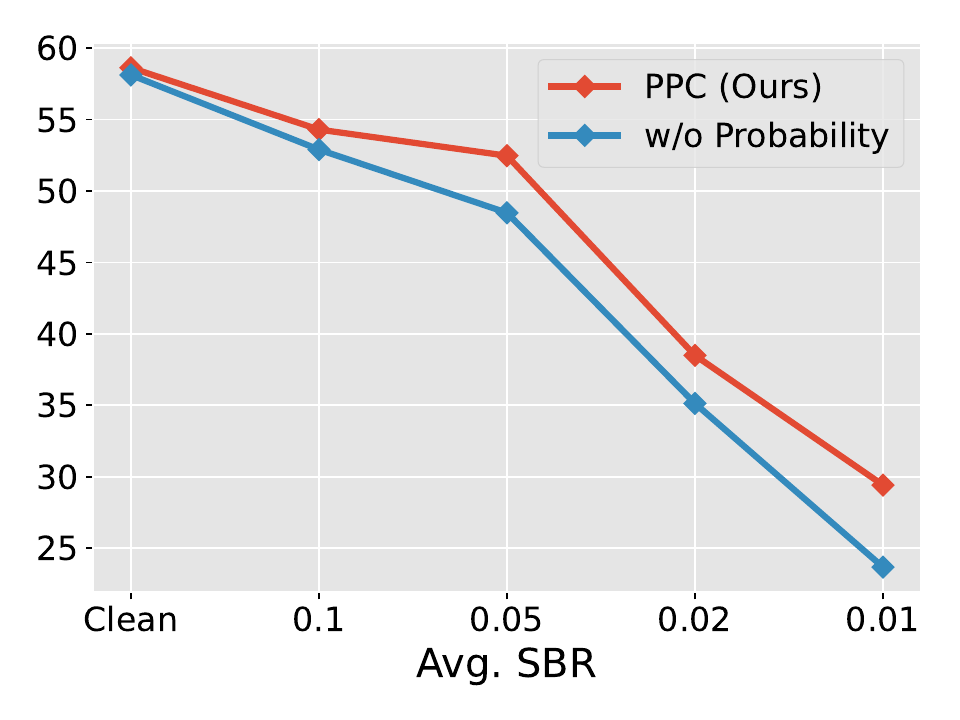}
\vspace{-15pt}
\caption{}
\label{fig:ablation_probs}
\end{subfigure}
\label{fig:ablation}
\vspace{-20pt}
\caption{\textbf{Ablation Study of PPC Components:} Performance of our approach (a) without FPPS and NPD Filtering, and (b) without probability attribute.}
\vspace{-5pt}
\end{figure}

First, we analyze the effectiveness of both NPD Filtering and FPPS individually. Fig.~\ref{fig:ablation_npdfps} shows the results of our approach without NPD and FPPS. NPD filtering shows a significant gain in performance, especially under very low SBR scenarios. FPPS shows an additional 2-4\% improvement in mAP under very low SBR conditions.

Second, we show the performance of our method on point clouds without the probability attribute in Fig.~\ref{fig:ablation_probs}. This is equivalent to using our approach with a conventional point cloud (\ie, all points with probability 1).
The probability attribute accounts for about a 4-5\% gain in mAP performance and is significant in very low SBR conditions.

\begin{figure}[htp!]
\begin{subfigure}{0.49\linewidth}
\includegraphics[width=\linewidth]{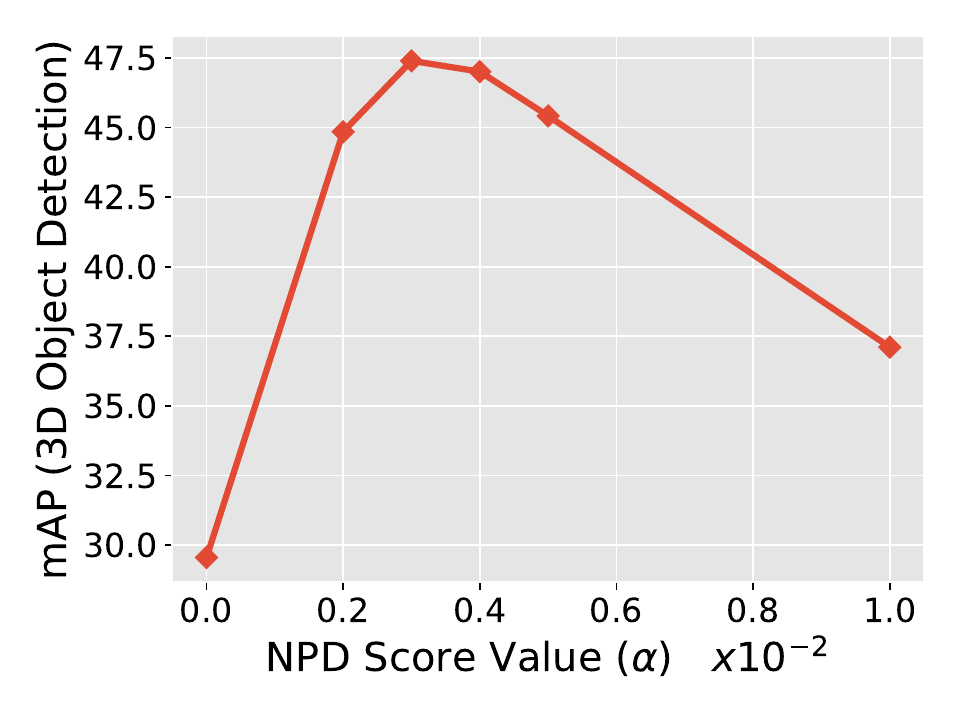}
\vspace{-15pt}
\caption{}
\label{fig:ablation_npdthreh}
\end{subfigure}
\begin{subfigure}{0.49\linewidth}
\includegraphics[width=\linewidth]{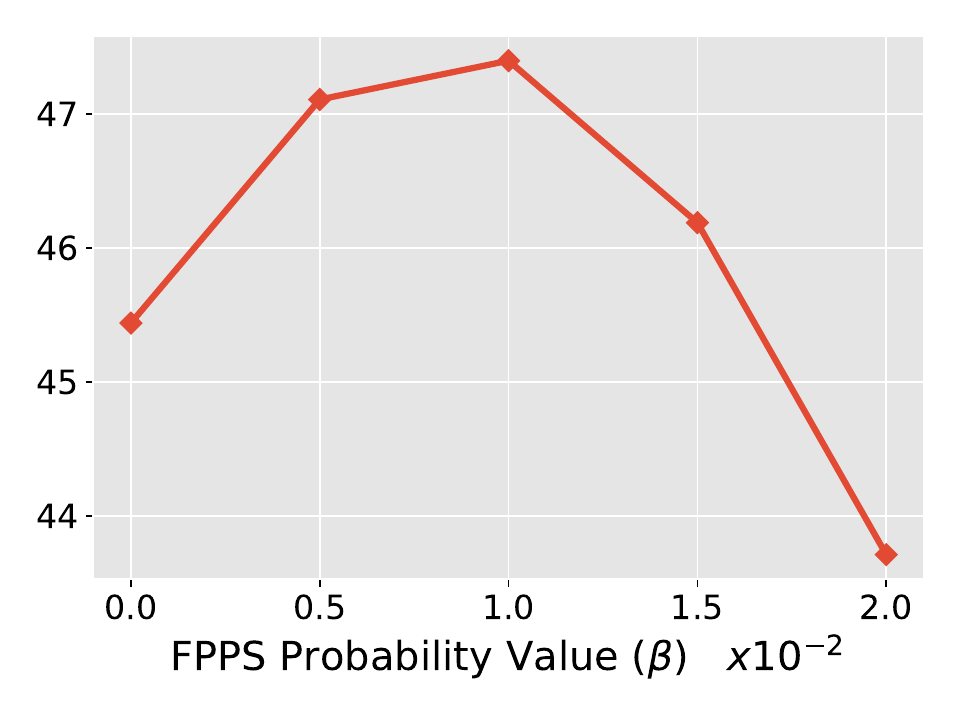}
\vspace{-15pt}
\caption{}
\label{fig:ablation_fppsthreh}
\end{subfigure}
\begin{subfigure}{0.49\linewidth}
\includegraphics[width=\linewidth]{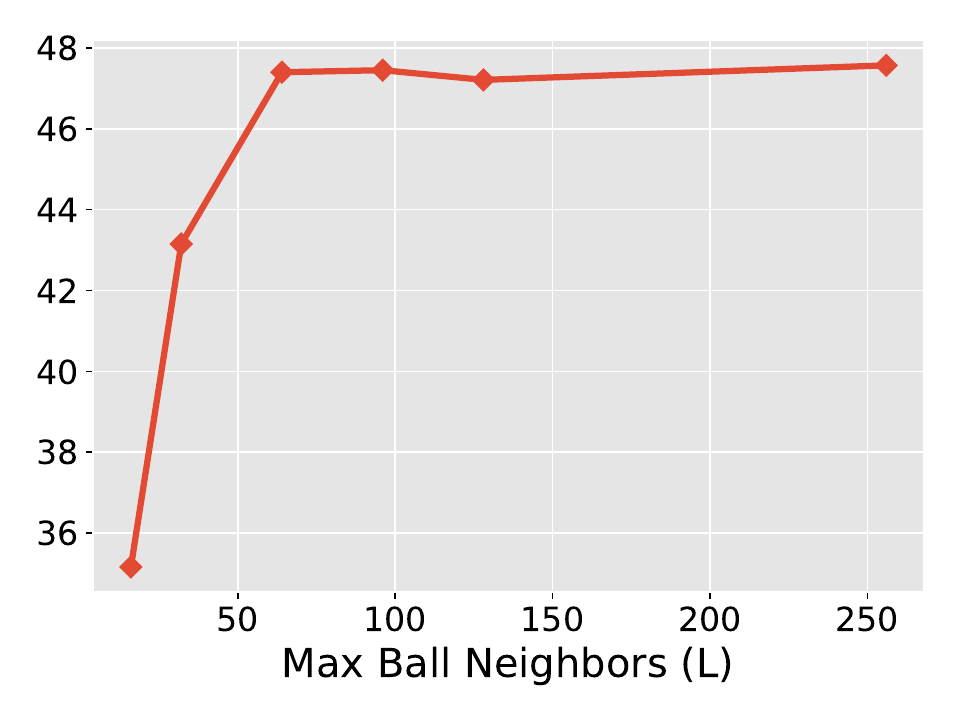}
\vspace{-15pt}
\caption{}
\label{fig:ablation_neighbors}
\end{subfigure}
\begin{subfigure}{0.49\linewidth}
\includegraphics[width=\linewidth]{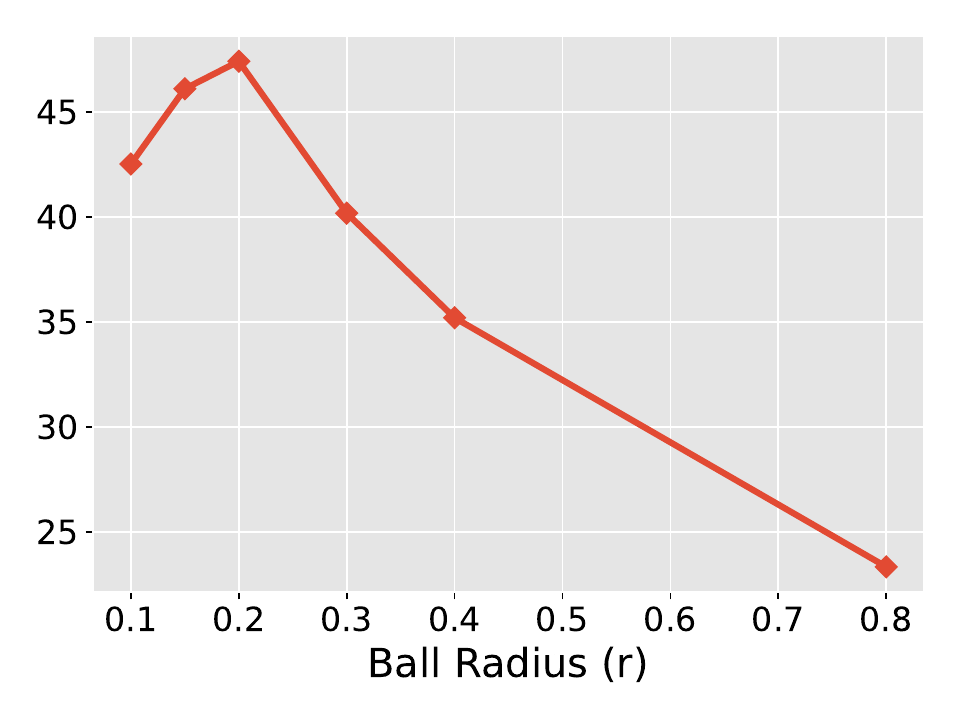}
\vspace{-15pt}
\caption{}
\label{fig:ablation_radius}
\end{subfigure}
\vspace{-5pt}
\caption{\textbf{Ablation Study of PPC Hyperparameters:} Performance of our approach with varying (a) NPD Score Value, (b) FPPS Value, (c) Max Ball Neighbors, and (d) Ball Radius.}
\label{fig:ablation_hyp}
\vspace{-10pt}
\end{figure}

We also show the performance of our approach by varying the hyperparameters of NPD filtering ($\alpha$) and FPPS ($\beta$). Fig.~\ref{fig:ablation_npdthreh} and \ref{fig:ablation_fppsthreh} show the mAP on the complete SUN RGB-D test set of all SBR levels. 
We chose the best performing value of $\alpha$ = 0.003 and $\beta$ = 0.01 for our models.

We also analyze our method by varying the hyperparameters of NPD score calculation, \ie, max ball neighbors ($L$) and ball radius ($r$). Fig.~\ref{fig:ablation_neighbors} and \ref{fig:ablation_radius} show the mAP on the complete SUN RGB-D test set of all SBR levels. 
We find an optimal NPD score value ($\alpha$) for each experiment.
Increasing the radius too much starts to hurt the performance, as noisy sparse points have more neighbors if the ball radius is bigger.
Performance improves as the value of $L$ increases, but saturates around 64.
We chose the values of $r$ = 0.2 and $L$ = 64 for our models.

Table~\ref{tab:ablation_inference_time} analyzes the total per-scene runtime of our method on the SUN RGB-D dataset. We use a single RTX 2070 Super GPU machine for inference time calculation. Adding FPPS adds no computational overhead, whereas adding NPD filtering adds less than 8\% of runtime with our implementation.

\begin{table}[H]
\centering
\footnotesize
\vspace{-5pt}
\begin{tabularx}{\linewidth}{Y|Y}
\toprule
& Inference Time (ms) \\
\midrule
PPC & 95 \\
PPC w/o FPPS & 95\\
PPC w/o NPD & 88\\
PPC w/o FPPS \& NPD & 88\\
\bottomrule
\end{tabularx}
\vspace{-5pt}
\caption{\textbf{Ablation Study of Runtime:} Our method adds no significant computational cost to runtime.}
\vspace{-5pt}
\label{tab:ablation_inference_time}
\end{table}

\begin{figure}[htp!]
\centering
\vspace{-5pt}
\includegraphics[width=0.5\linewidth]{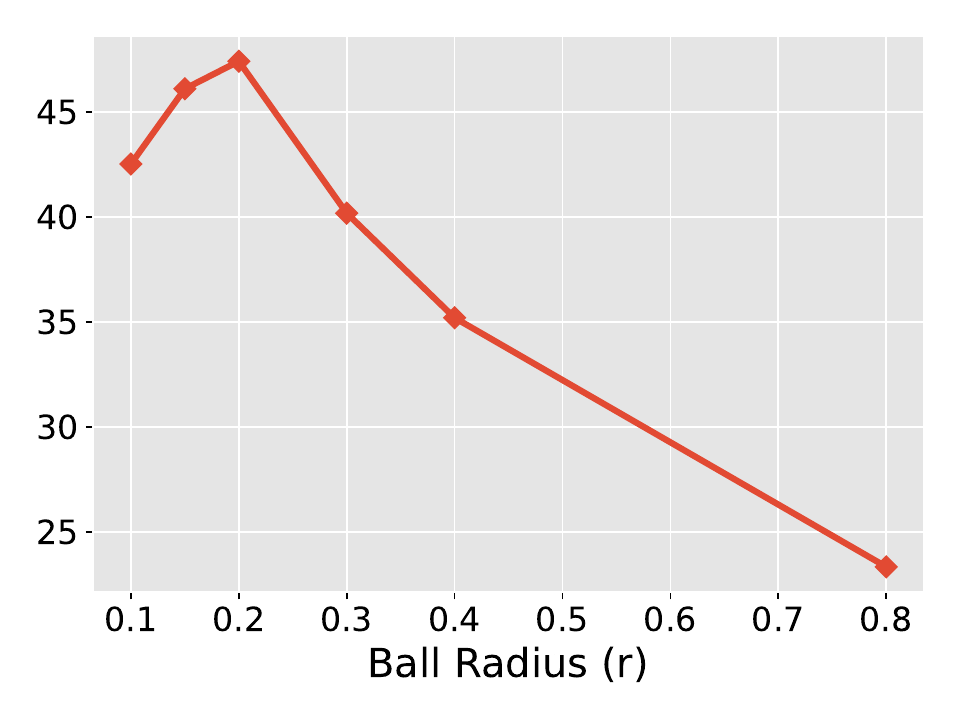}
\vspace{-5pt}
\caption{\textbf{Ablation Study of Thresholding baseline:} Performance of the Thresholding baseline with varying threshold used for the model.}
\label{fig:ablation_thresh}
\vspace{-5pt}
\end{figure}

We analyze the performance of the Thresholding baseline by varying the threshold used for the model. Fig.~\ref{fig:ablation_thresh} shows AP@25 results on the complete SUN RGB-D test set of all SBR levels. We select the best performing threshold (=1.1) for evaluating this baseline.

\clearpage
\clearpage

\section{Additional 3D Object Detection Results}
In this section, we include implementation details, additional results, and analysis that supplement the experiments in the main paper.

\begin{table*}[tp!]
\footnotesize
\centering
\setlength\tabcolsep{18pt}
\begin{tabular}{c  c  c  c  c  c  c}
\toprule
 & Matched &  Thresholding &  PointClean &  Score & PathNet & \textbf{PPC}\\
Category &  Filtering &  &  Net~\cite{rakotosaona2020pointcleannet}&  Denoise~\cite{luo2021score} & ~\cite{wei2024pathnet} &  \textbf{(Ours)} \\
\midrule
Bed & 54.37 & 67.33 & 53.59 & 68.57& 67.53 & \textbf{72.97}\\
Sofa & 16.20 & 28.15 & 22.00 & 38.25 & 38.03 & \textbf{45.19}\\
Table & 29.61 & 37.99 & 30.20 & 33.93 & 32.17 & \textbf{40.47}\\
Bathtub &  6.04 & 25.14 & 2.71 & 14.04 & 13.46 & \textbf{54.32}\\
Desk & 7.97 & 14.16 & 7.78 & 8.33 & 9.18 & \textbf{17.37}\\
Bookshelf & 2.12 & 4.85 & 0.67 & 1.01 & 0.89 & \textbf{9.80}\\
\midrule
Chair  & 22.05 & 34.27 & 22.92 & 27.45 & 25.43 & \textbf{47.47}\\
Night Stand & 3.10 & 14.45 & 7.41 & 10.23 & 9.13 & \textbf{30.49}\\
Dresser & 2.58 & 4.64 & 2.80 & 2.44 & 2.09 & \textbf{13.73}\\
\bottomrule
\end{tabular}
\vspace{-5pt}
\caption{\textbf{Category-wise 3D Object Detection Results}: Table shows per category AP@25 results on the SUN RGB-D dataset under low SBR (0.02) conditions. Our approach outperforms all baselines and shows large gains for smaller object categories (below the line) like chairs and nightstands.}
\label{tab:supp3ddetectionresults}
\end{table*}

\begin{table*}
\centering
\setlength\tabcolsep{11pt} 
\footnotesize
\begin{tabular}{c ccc c ccc c ccc c ccc c ccc}
\toprule
Avg. SBR & \multicolumn{3}{c}{Car} && \multicolumn{3}{c}{Pedestrian} && \multicolumn{3}{c}{Cyclist} \\
\cline{2-4}\cline{6-8}\cline{10-12}
 & Easy & Mod & Hard & & Easy & Mod & Hard & & Easy & Mod & Hard \\
\midrule
Matched Filtering & 79.02 & 59.95 & 57.67 && 50.85 & 47.06 & 43.51 && 68.78 & 43.74 & 41.10 \\
Thresholding & 78.73 & 59.40 & 55.35 && 54.45 & 49.23 & 45.38 && 68.13 & 44.96 & 42.64 \\
\textbf{PPC (Ours)} & \textbf{79.10} & \textbf{60.29} & \textbf{59.08} && \textbf{60.42} & \textbf{55.39} & \textbf{50.82} && \textbf{71.99} & \textbf{47.76} & \textbf{44.84} \\
\bottomrule
\end{tabular}
\vspace{-5pt}
\caption{\textbf{KITTI 3D Detection Comparison}: Table shows mAP for easy, moderate, and hard difficulty levels on KITTI val split calculated with 11 recall positions for PV-RCNN architecture under low SBR (0.01) conditions.}
\label{tab:supp3ddetectionresults_kittilevels}
\end{table*}

\subsection{Datasets}
We evaluate our approach on 3D object detection benchmarks of SUN RGB-D~\cite{song2015sun} and KITTI~\cite{geiger2012we}.
SUN RGB-D consists of $\sim$10K RGB-D scans of \textit{indoor} scenes annotated with 3D bounding boxes. 
The dataset also provides camera intrinsic and extrinsic parameters to convert depth scans to 3D point clouds.
We use the standard evaluation protocol that considers 10 common object categories.
KITTI dataset is a widely used \textit{outdoor} autonomous driving dataset containing $\sim$7.4k annotated scenes with LiDAR point clouds.
We follow the standard evaluation protocol using three categories: car, pedestrian, and cyclist.

\subsection{Implementation Details} 
We implement our method using the MMDetection3D framework~\cite{mmdet3d2020} provided by OpenMMLab and use the same evaluation procedure as the previous literature.
For the SUN RGB-D dataset, the color information from the point clouds is not used for inference and is only used for visualization. For the KITTI dataset, the reflection intensity information from LiDAR point clouds is used as input for all methods.

\subsection{Category-wise Performance}
Table~\ref{tab:supp3ddetectionresults} shows per-category AP@25 results on the SUN RGB-D dataset for all methods under low SBR (0.02) conditions. Our approach shows significant gains for all categories, particularly larger gains for \emph{small-sized} object categories (\eg chair, nightstand, dresser) which typically suffer the most under low SBR conditions.
Table~\ref{tab:supp3ddetectionresults_kittilevels} shows mAP on the KITTI val split for PV-RCNN architecture under low SBR (0.01) conditions, calculated with 11 recall positions in a standard format similar to previous works. PPC outperforms the baselines in all categories.

\subsection{Visualizations and Observations}
\begin{figure*}
\centering
\includegraphics[width=\linewidth]{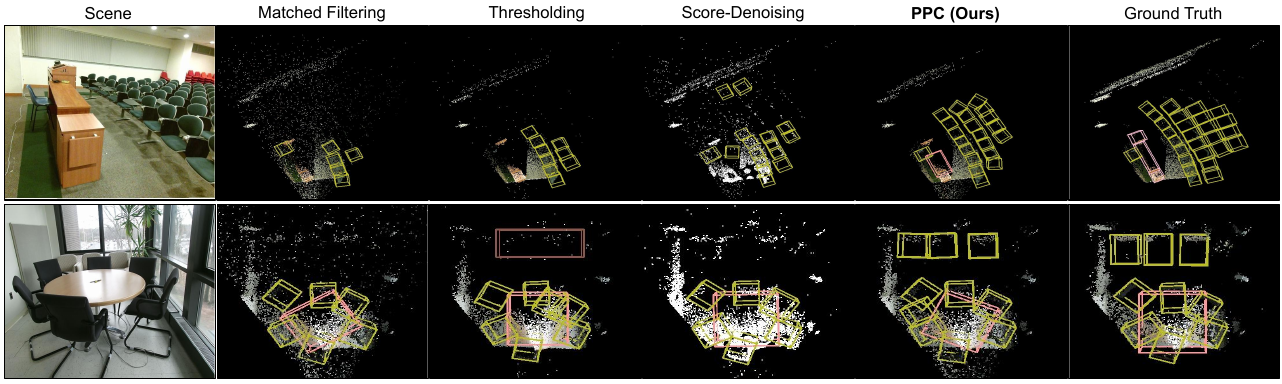}
\vspace{-15pt}
\caption{\textbf{3D Object Detection Results:} Figure shows scenes from the SUN RGB-D dataset under medium SBR (0.1) conditions. The first scene contains multiple rows of small objects (\textcolor{yellow}{chair}). Baselines fail to detect farther rows of chairs. Our approach detects most \textcolor{yellow}{chairs} and \textcolor{Lavender}{table}.}
\label{fig:results_5_50_supp}
\end{figure*}

\begin{figure*}
\centering
\includegraphics[width=\linewidth]{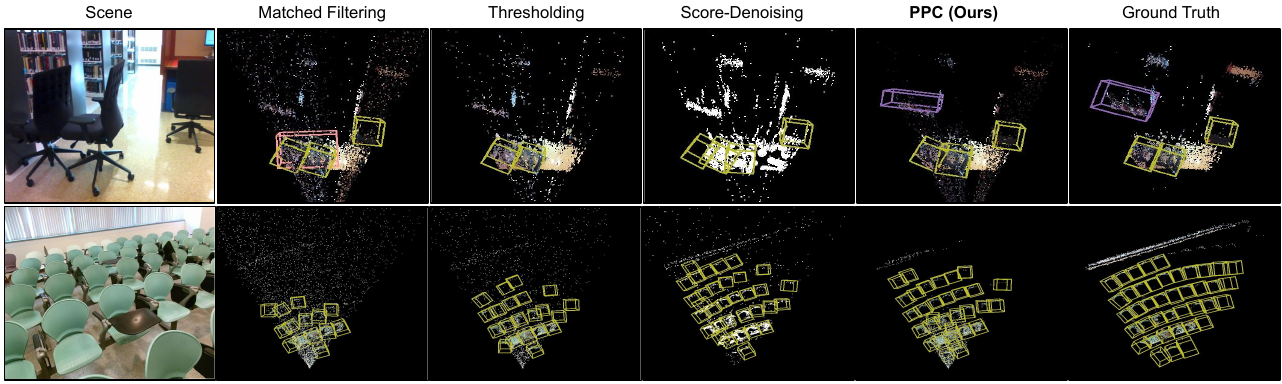}
\vspace{-15pt}
\caption{\textbf{3D Object Detection Results:} Figure shows scenes from the SUN RGB-D dataset under medium SBR (0.05) conditions. Scenes consist of small (\textcolor{yellow}{chair}) and farther objects (\textcolor{blue-violet}{bookshelf}). Our approach detects most objects with no false detections.}
\label{fig:results_5_100_supp}
\end{figure*}

\begin{figure*}
\centering
\includegraphics[width=\linewidth]{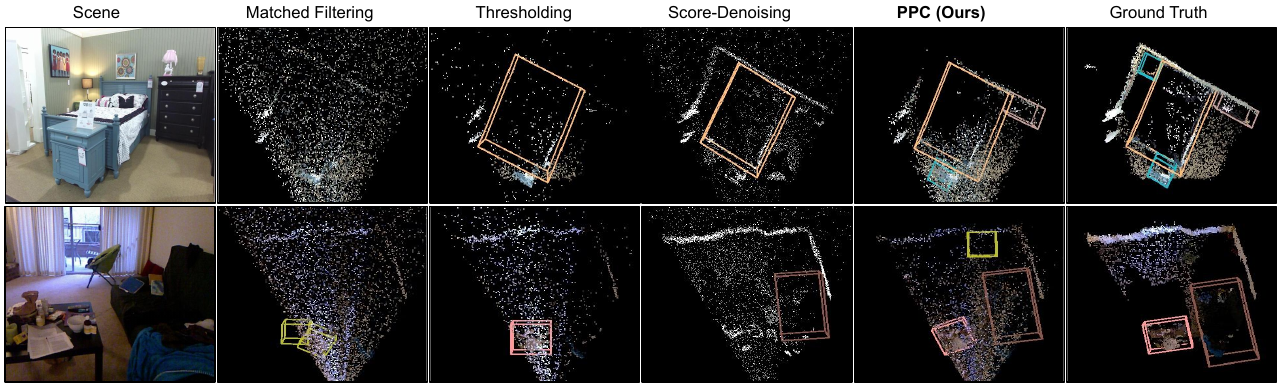}
\vspace{-15pt}
\caption{\textbf{3D Object Detection Results:} Figure shows scenes from the SUN RGB-D dataset under low SBR (0.02) conditions. Scenes consist of small (\textcolor{cyan}{nightstand}) and occluded objects (\textcolor{Lavender}{table}). Our approach performs better than all baselines.}
\label{fig:results_1_50_supp}
\end{figure*}

\begin{figure*}
\vspace{-5pt}
\centering
\includegraphics[width=\linewidth]{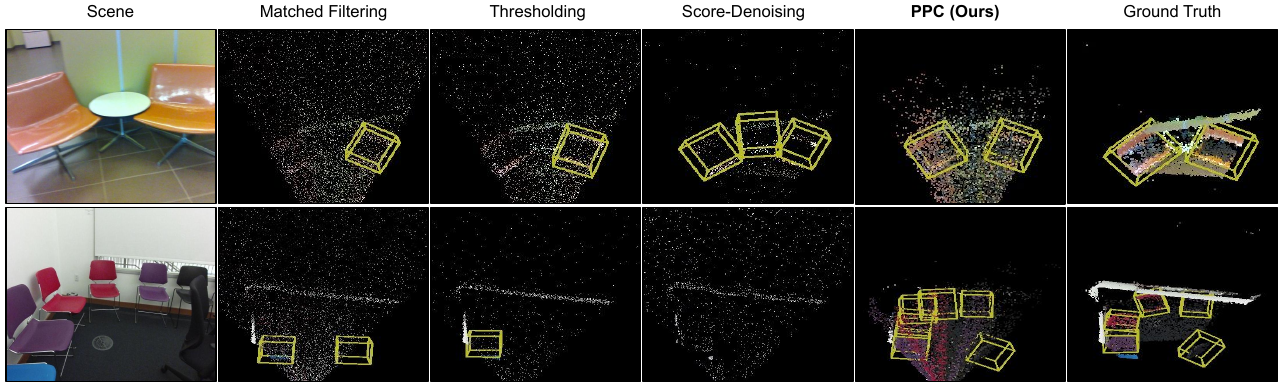}
\vspace{-15pt}
\caption{\textbf{3D Object Detection Results:} 
The figure shows scenes from the SUN RGB-D dataset under low SBR (0.01) conditions.
Baselines fail to detect numerous objects (\textcolor{yellow}{chair}) due to noise, whereas our approach detects most objects in the scene.}
\label{fig:results_1_100_supp}
\end{figure*}

\begin{figure*}
\centering
\includegraphics[width=\linewidth]{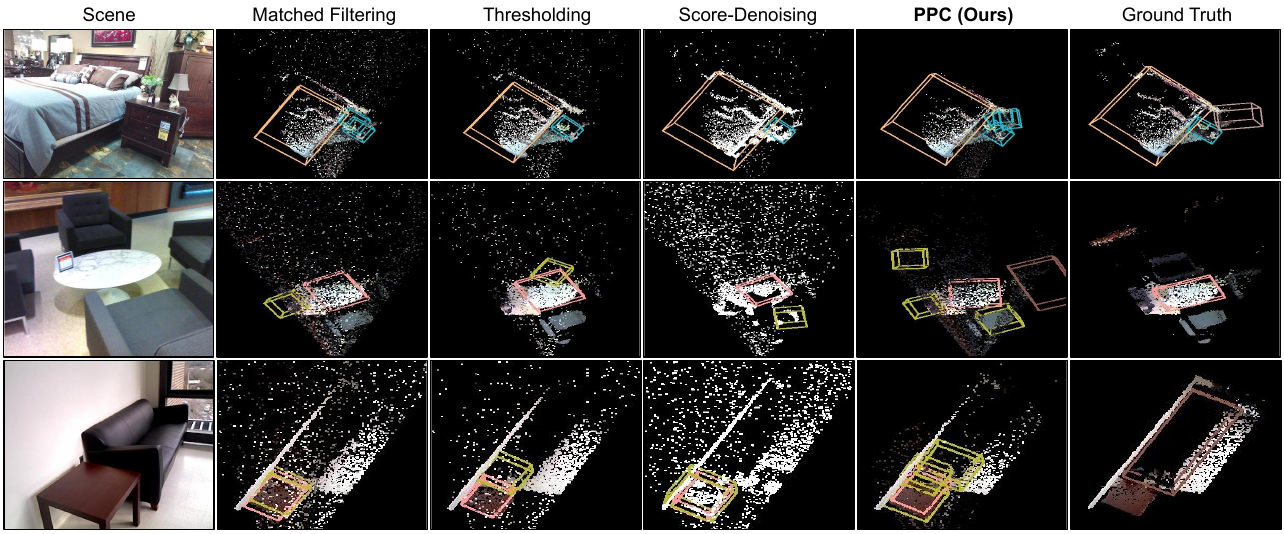}
\vspace{-15pt}
\caption{\textbf{3D Detection Failure Cases:} The first scene shows a scenario where PPC detects multiple boxes for the same object (\textcolor{cyan}{nightstand}). The second and third scenes show scenarios where a couch is detected as a chair by PPC and the baselines. Single-sitter couches are often detected as chairs by this model.}
\label{fig:supp_results_failure}
\end{figure*}

\begin{figure*}
\centering
\includegraphics[width=\linewidth]{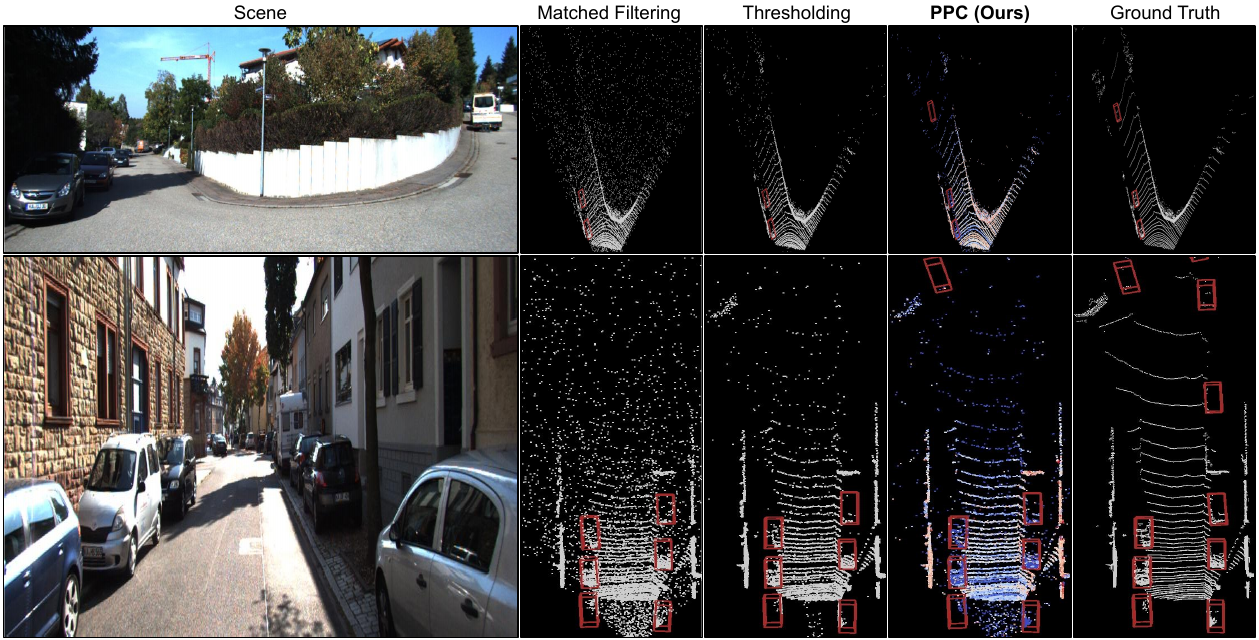}
\vspace{-20pt}
\caption{\textbf{3D Object Detection Results:} Figure shows scenes from the KITTI dataset under medium SBR (0.05) conditions. Baselines fail to detect farther \textcolor{red}{cars}. PPC is more robust for distant objects.}
\label{fig:kitti_results_5_100_supp}
\end{figure*}

\begin{figure*}
\centering
\includegraphics[width=\linewidth]{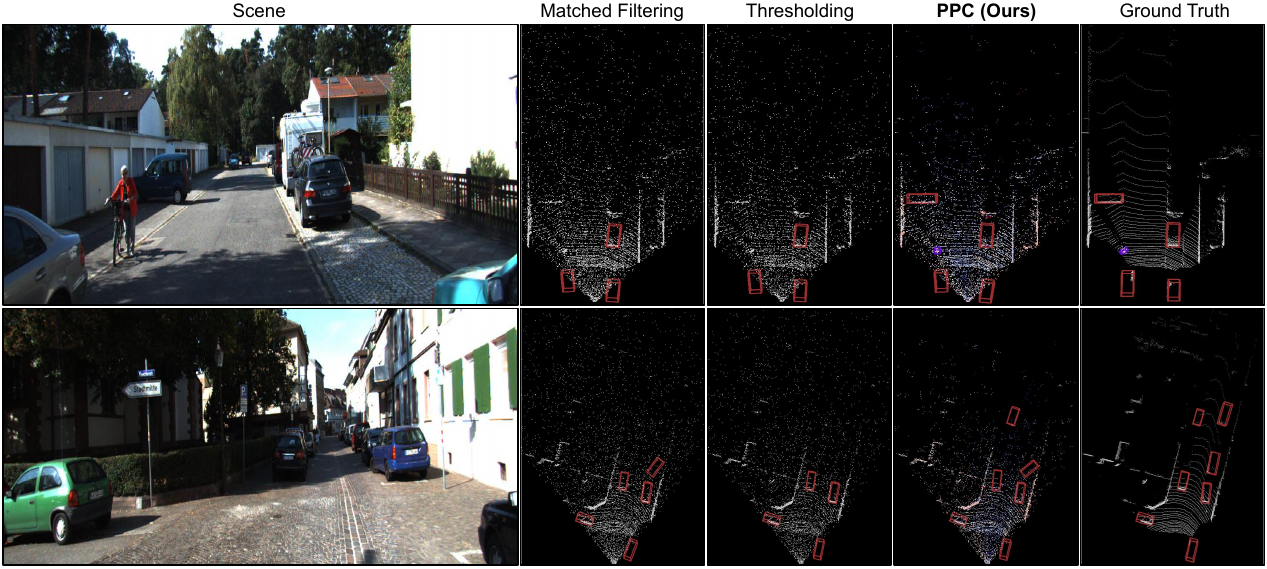}
\vspace{-20pt}
\caption{\textbf{3D Object Detection Results:} Figure shows scenes from the KITTI dataset under low SBR (0.02) conditions. The first scene shows a scenario where baselines fail to detect objects like farther \textcolor{red}{cars} and the \textcolor{blue-violet}{pedestrian}. PPC is more robust for distant objects.
}
\label{fig:kitti_results_5_250_supp}
\end{figure*}

\begin{figure*}
\centering
\includegraphics[width=\linewidth]{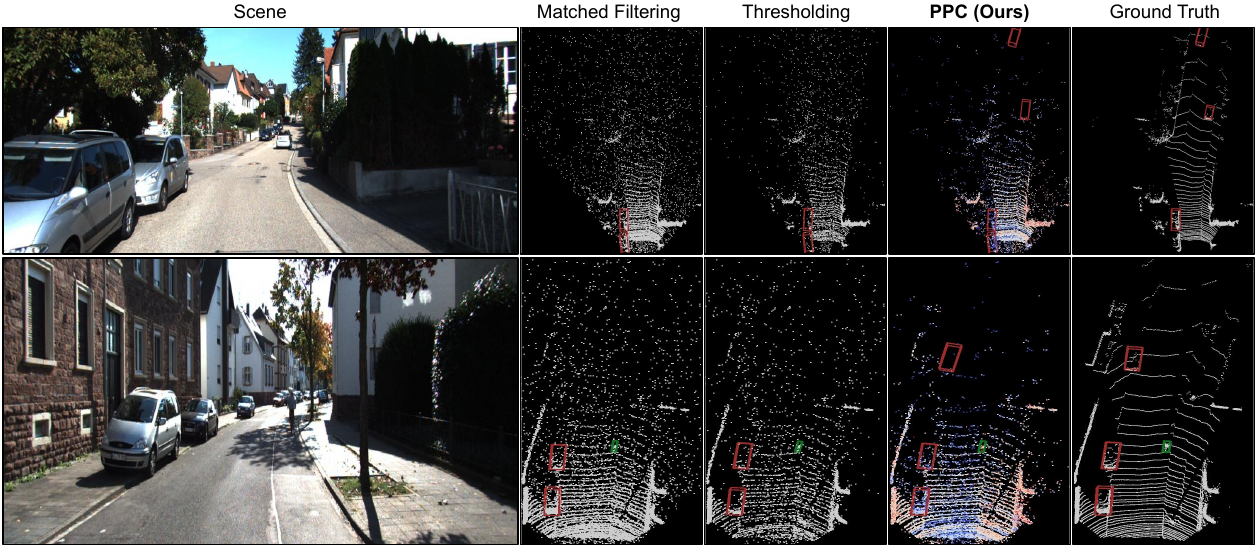}
\vspace{-20pt}
\caption{\textbf{3D Object Detection Results:} Figure shows scenes from the KITTI dataset under low SBR (0.01) conditions. Baselines struggle with farther \textcolor{red}{cars}, whereas PPC detects most objects.}
\label{fig:kitti_results_5_500_supp}
\end{figure*}

\begin{figure*}
\centering
\includegraphics[width=\linewidth]{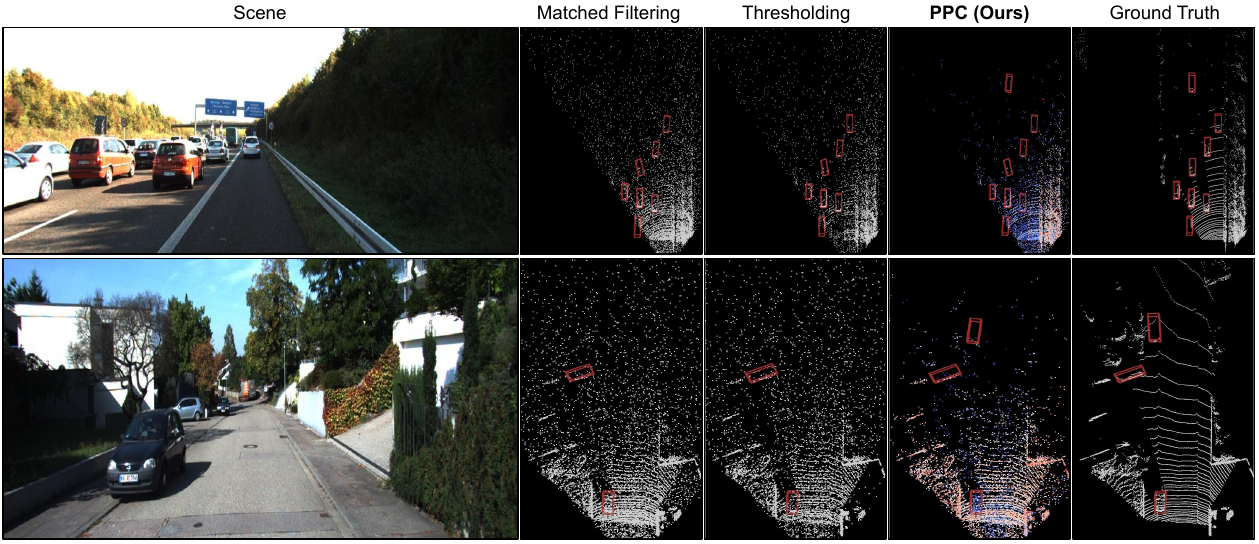}
\vspace{-20pt}
\caption{\textbf{3D Object Detection Results:} Figure shows scenes from the KITTI dataset under low SBR (0.005) conditions. Baselines fail to detect dark and farther objects like black \textcolor{red}{cars}. PPC is more robust for distant objects.}
\label{fig:kitti_results_5_1000_supp}
\end{figure*}

Fig.~\ref{fig:results_5_50_supp} to \ref{fig:results_1_100_supp}
show visualizations of 3D object detection on the SUN RGB-D dataset for all methods under different SBR conditions using the VoteNet architecture.
Fig.~\ref{fig:results_5_50_supp} and \ref{fig:results_5_100_supp} show complex scenes with a large number of small objects (\eg chairs).
Baselines fail to detect a lot of small and farther objects (last row of chairs), whereas PPC detects the most objects accurately.
Fig.~\ref{fig:results_1_50_supp} and \ref{fig:results_1_100_supp} show scenes with very low SBR conditions.
Baselines fail to detect many objects, whereas our approach performs significantly better even in the presence of a large amount of noise.
Fig.~\ref{fig:supp_results_failure} shows a few failure cases for our method. PPC can sometimes detect multiple overlapping bounding boxes for the same object under noise.
Couch or single-sitter couches are often detected as chairs by PPC or other baselines. Fig.~\ref{fig:kitti_results_5_100_supp} to \ref{fig:kitti_results_5_1000_supp} show scenes from the KITTI val dataset under varying SBR conditions using PV-RCNN architecture.
Baselines fail to detect many objects, like farther cars and pedestrians, whereas our approach detects most objects.

\clearpage
\clearpage

\section{More 3D Detection Architectures} 
\begin{table*}[b!]
\footnotesize
\centering
\begin{tabular}{c ccc ccc ccc ccc cc}
\toprule
Avg. SBR & \multicolumn{2}{c}{Clean} && \multicolumn{2}{c}{0.1} && \multicolumn{2}{c}{0.05} && \multicolumn{2}{c}{ 0.02 } && \multicolumn{2}{c}{0.01} \\
\cline{2-3}\cline{5-6}\cline{8-9}\cline{11-12}\cline{14-15}
 & \scriptsize AP@25 & \scriptsize AP@50 && \scriptsize AP@25 &\scriptsize AP@50 &&\scriptsize  AP@25 &\scriptsize AP@50 &&\scriptsize AP@25 &\scriptsize AP@50 &&\scriptsize AP@25 &\scriptsize AP@50\\
\midrule
Matched Filtering & 63.37 & 35.51 && 53.89 & 27.64 && 53.23 & 24.67 && 37.54 & 10.99 && 33.17 & 7.98\\
Thresholding & 64.25 & 36.17 && 59.57 & 33.44 && 58.82 & 32.45 && 42.43 & 18.17 && 39.51 &  12.61 \\
\textbf{PPC (Ours)} & \textbf{64.36} & \textbf{36.94} && \textbf{61.51} & \textbf{35.69} && \textbf{60.19} & \textbf{31.38} && \textbf{53.21} & \textbf{25.37} && \textbf{46.84} & \textbf{20.14} \\
\bottomrule
\end{tabular}
\vspace{-5pt}
\caption{\textbf{3D Detection Comparison using camera-LiDAR fusion architecture}: AP@0.25 and AP@0.50 results on the SUN RGB-D dataset using ImVoteNet show significant gains for PPC for all SBR levels.}
\label{tab:supp3ddetectionresults_sunrgbdfusion}
\end{table*}

\begin{table*}[b!]
\vspace{10pt}
\footnotesize
\centering
\begin{tabular}{c ccc ccc ccc ccc cc}
\toprule
Avg. SBR & \multicolumn{2}{c}{Clean} && \multicolumn{2}{c}{0.1} && \multicolumn{2}{c}{0.05} && \multicolumn{2}{c}{ 0.02 } && \multicolumn{2}{c}{0.01} \\
\cline{2-3}\cline{5-6}\cline{8-9}\cline{11-12}\cline{14-15}
 & \scriptsize AP@25 & \scriptsize AP@50 && \scriptsize AP@25 &\scriptsize AP@50 &&\scriptsize  AP@25 &\scriptsize AP@50 &&\scriptsize AP@25 &\scriptsize AP@50 &&\scriptsize AP@25 &\scriptsize AP@50\\
\midrule
Matched Filter & 64.98 & 48.28 && 61.52 & 45.09 && 60.82 & 43.92 && 51.12 & 34.09 && 45.29 & 27.97 \\
Thresholding & 64.50 & 47.94 && 61.08 & 44.71 && 61.19 & 44.29 && 51.97 & 34.87 && 48.13 & 28.64 \\
\textbf{PPC (Ours)} & \textbf{65.53} & \textbf{49.35} && \textbf{62.58} & \textbf{46.71} && \textbf{61.98} & \textbf{48.28} &&  \textbf{56.46} & \textbf{38.03} && \textbf{51.21} & \textbf{31.16}  \\
\bottomrule
\end{tabular}
\vspace{-5pt}
\caption{\textbf{3D Detection Comparison using LiDAR-only tranformer-based architecture}: AP@0.25 and AP@0.50 results on the SUN RGB-D dataset using Uni3DETR show significant gains for PPC under low SBR conditions.}
\label{tab:supp3ddetectionresults_uni3DETR}
\end{table*}

\begin{table*}[b!]
\vspace{10pt}
\centering
\footnotesize
\setlength\tabcolsep{4pt} 
\begin{tabular}{c ccc c ccc c ccc c ccc c ccc}
\toprule
Avg. SBR & \multicolumn{3}{c}{Clean} && \multicolumn{3}{c}{0.05} && \multicolumn{3}{c}{0.02} && \multicolumn{3}{c}{0.01} && \multicolumn{3}{c}{0.005} \\
\cline{2-4}\cline{6-8}\cline{10-12}\cline{14-16}\cline{18-20}
 & Car & Ped & Cyc &  & Car & Ped & Cyc & & Car & Ped & Cyc & & Car & Ped & Cyc & & Car & Ped & Cyc  \\
\midrule
Matched Filtering & 77.08 & \textbf{52.78} & 64.49 && 68.25 & 49.52 & 58.96 && 64.13 & 47.67 & 46.45 && 54.43 & 41.61 & 41.76 && 45.03 & 32.46 & 31.06\\
Thresholding & \textbf{77.34} & 52.09 & 64.81 && 68.06 & 49.63 & 59.09 && 63.87 & 47.92 & 46.96 && 54.18 & 40.88 & 42.18 && 45.11 & 32.79 & 31.89 \\
\textbf{PPC (Ours)} & 77.19 & 52.12 & \textbf{65.21} && \textbf{69.12} & \textbf{50.23} & \textbf{62.44} && \textbf{65.63} & \textbf{49.27} & \textbf{48.09} && \textbf{56.39} & \textbf{45.77} & \textbf{44.46} && \textbf{47.24} & \textbf{38.74} & \textbf{34.89} \\
\bottomrule
\end{tabular}
\vspace{-5pt}
\caption{\textbf{3D Detection Comparison using LiDAR-only pillar-based architecture}: mAP results for car, pedestrian, and cyclist categories on moderate difficulty of KITTI val split calculated with 11 recall positions for PointPillars. Our method shows significant gains under low SBR conditions.}
\label{tab:supp3ddetectionresults_kittipillar}
\end{table*}

In this section, we evaluate our PPC approach using a variety of 3D object detection architectures. First, we evaluate a camera-LiDAR fusion approach, ImVoteNet \cite{qi2020imvotenet}. Table~\ref{tab:supp3ddetectionresults_sunrgbdfusion} includes the comparison on the SUN RGB-D dataset, which shows significant improvement for all SBR levels. Second, we evaluate using a recent LiDAR-only transformer-based architecture Uni3DETR~\cite{NEURIPS2023_7d60bfd8}.
Table~\ref{tab:supp3ddetectionresults_uni3DETR} includes mAP comparison on the SUN RGB-D dataset, which shows performance improvement for low SBR levels. Lastly, we evaluate using a Pillar-based architecture, PointPillars~\cite{lang2019pointpillars}. Table~\ref{tab:supp3ddetectionresults_kittipillar} includes mAP for car, pedestrian, and cyclist categories on moderate difficulty of KITTI val split, calculated with 11 recall positions. Our method shows significant improvements under low SBR conditions for pedestrian and cyclist categories.

PPC shows significant gains under low SBR for all detection architectures, which shows its versatility for a wide range of 3D detection models. The gain is large for point-net or transformer-based architectures (Uni3DETR, VoteNet, and ImVoteNet) as they suffer the most from the low SBR noise. The gain is significant but comparatively smaller for voxel or pillar-based architectures (PointPillars and PV-RCNN). Intuitively, this could be because the spurious points with large depth errors are away from the surface, and are not part of the same voxel or pillar as the surface points. Hence, their impact on the performance is also smaller.

\subsection{Additional 3D inference tasks}
Our approach is easy to extend to other 3D inference tasks, \eg.~point cloud classification and point segmentation, with minimal modifications. Table below shows preliminary results for point cloud classification using PointNext \cite{qian2022pointnext} backbone on ScanObjectNN \cite{uy-scanobjectnn-iccv19} dateset. PPC outperforms the baselines under low SBR conditions.

\begin{table}[ht!]
\centering
\footnotesize
\begin{tabularx}{\linewidth}{Y Y Y}
\toprule
Avg. SBR & 0.1 & 0.01\\
\midrule
Matched Filtering & 62.3 & 49.8 \\
Threshoding & 64.4 & 53.4 \\
PPC (Ours) & 70.5 & 58.4\\
\bottomrule
\end{tabularx}
\vspace{-5pt}
\caption{Classification overall accuracy on ScanObjectNN dataset.}
\vspace{-5pt}
\end{table}

\clearpage
\clearpage

\section{Comparison with Denoised 3D Temporal Histograms}
An effective approach for removing noise in 3D sensing systems described in this work is to denoise the 3D temporal histograms.
Current state-of-the-art denoising methods for temporal histograms denoising~\cite{peng2020photon} show high performance on depth reconstruction tasks.
Hence, we also evaluate our approach and baselines using denoised temporal histograms.
We use a 3D-CNN denoising model~\cite{peng2020photon} to denoise the temporal bins, which are then used to construct point clouds for inference.
We compare our method with the baselines under low SBR (0.02) conditions in Table~\ref{tab:other_approaches}.
As expected, all methods perform better with denoised temporal histograms.
Our method shows a further gain in AP@0.25 of about 3-4\%, which shows that 3D inference can benefit from our PPC approach with denoised temporal histograms as well.

\begin{table}[htp!]
\centering
\footnotesize
\setlength\tabcolsep{2pt} 
\begin{tabular}{c c c c c c}
\toprule
Histogram& Thresholding & PointClean & Score-& PathNet & \textbf{PPC} \\
Denoising & & Net & Denoising && \textbf{(Ours)} \\
Method & &  & &&  \\
\midrule
- & 29.58 & 18.24 & 26.35 & 25.45 & \textbf{38.49} \\
3D-CNN ~\cite{peng2020photon} & 50.30 & 51.03 & 50.85 & 51.07 & \textbf{54.25} \\
Gaussian Filter & 38.79 & 40.12 & 43.36 & 43.25 & \textbf{50.93} \\
\bottomrule
\end{tabular}
\caption{\textbf{3D Object Detection Results}: Comparison of AP@0.25 results using denoised temporal histograms.}
\label{tab:other_approaches}
\end{table}

We also compare the effectiveness of our method with a non-learning-based histogram denoising method.
We use a 5x5 Gaussian filter in the spatial dimension to denoise histograms.
Matched filtering is still used over the temporal dimension. 
Table~\ref{tab:other_approaches} shows that our method has significant gains with Gaussian denoised histograms, and the performance is comparable to the results with 3D-CNN denoising.

\medskip
\noindent \textbf{Should we denoise 3D temporal histograms for inference?} Denoising methods like \cite{peng2020photon} require compute and memory-intensive 3D-CNN operations, which makes it infeasible for real-time applications.
It is thus not suitable for sensor on-chip processing.
It also requires a read-out of full 3D temporal histograms, which has a significantly high data-bandwidth cost (compared to only reading out a point cloud as considered in earlier approaches in the main paper).
However, it is an effective approach for denoising in non-real-time applications.

The Gaussian filter is a computationally cheap, non-learning-based operation and is feasible with sensor on-chip processing.
Table~\ref{tab:supp_inference_time} shows the inference time of our complete recognition pipeline (including denoising temporal histograms, point cloud processing, and inference) with different histogram denoising methods that are discussed. This suggests that, given a computational budget, a simple histogram denoising approach like a Gaussian filter is a good candidate for 3D recognition.


\begin{table}[htp!]
\centering
\footnotesize
\setlength\tabcolsep{15pt} 
\begin{tabular}{c c }
\toprule
Histogram Denoising Method & Runtime (ms) \\
\midrule
- & 95\\
3D-CNN & 7200 \\
Gaussian Filter & 98 \\
\bottomrule
\end{tabular}
\caption{\textbf{Runtime Time for 3D Detection:} Comparison of per-scene runtime time of our method using different histogram denoising methods.}
\label{tab:supp_inference_time}
\end{table}




\subsection{Comparison with Compressed 3D Timing Histograms}

Recently, compression techniques have been proposed to read out compressed representations~\cite{gutierrez2022compressive} of the temporal histograms to reduce data bandwidth requirements.
We also show the performance of our approach on such decompressed histograms in Table~\ref{tab:compressedhistograms}.
We use a lightweight Truncated Fourier (k=32) representation from ~\cite{gutierrez2022compressive} for evaluation.
Our approach is effective even under the data loss incurred due to compression and shows significant gains over the Thresholding baseline for 3D detection.

\begin{table}[htp!]
\centering
\footnotesize
\begin{tabular}{c c c}
\toprule
& Threshoding & \textbf{PPC (Ours)} \\
\midrule
Decompressed Histograms & 16.50 & \textbf{29.77}\\
\bottomrule
\end{tabular}
\caption{\textbf{3D Object Detection Results}: Comparison of AP@0.25 results using decompressed histograms.}
\label{tab:compressedhistograms}
\end{table}


\clearpage

\section{Point Clouds Under Extremely Low SBR Conditions}
\begin{figure}[ht!]
\centering
\begin{subfigure}[justification=centering]{0.48\linewidth}
\frame{
\includegraphics[width=\linewidth]{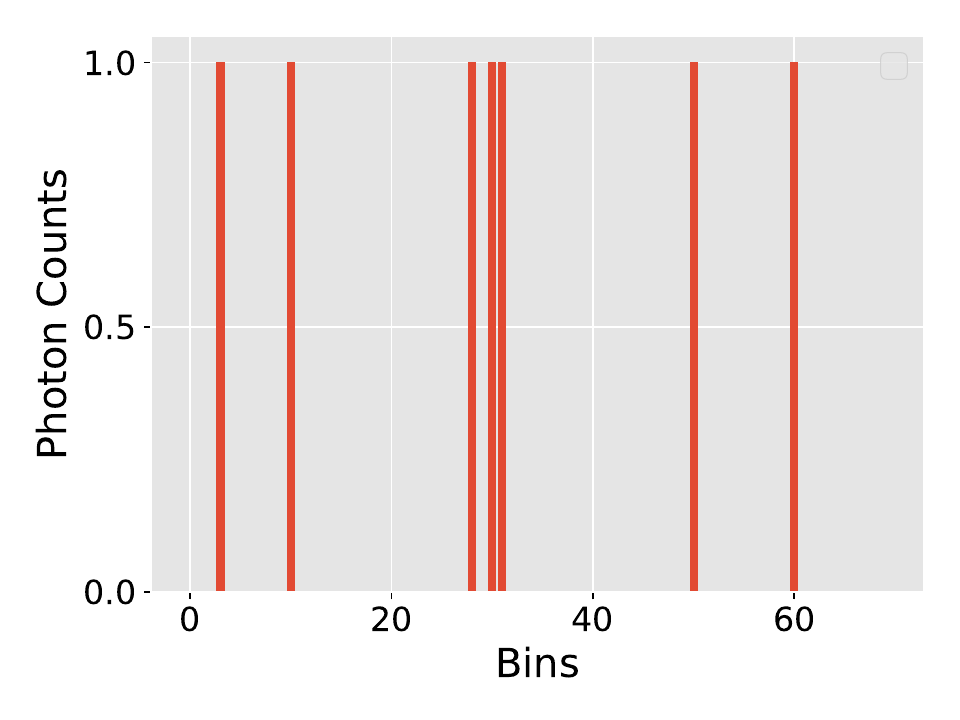}
}
\caption{Raw Histogram}
\end{subfigure}
\begin{subfigure}[justification=centering]{0.48\linewidth}
\frame{
\includegraphics[width=\linewidth]{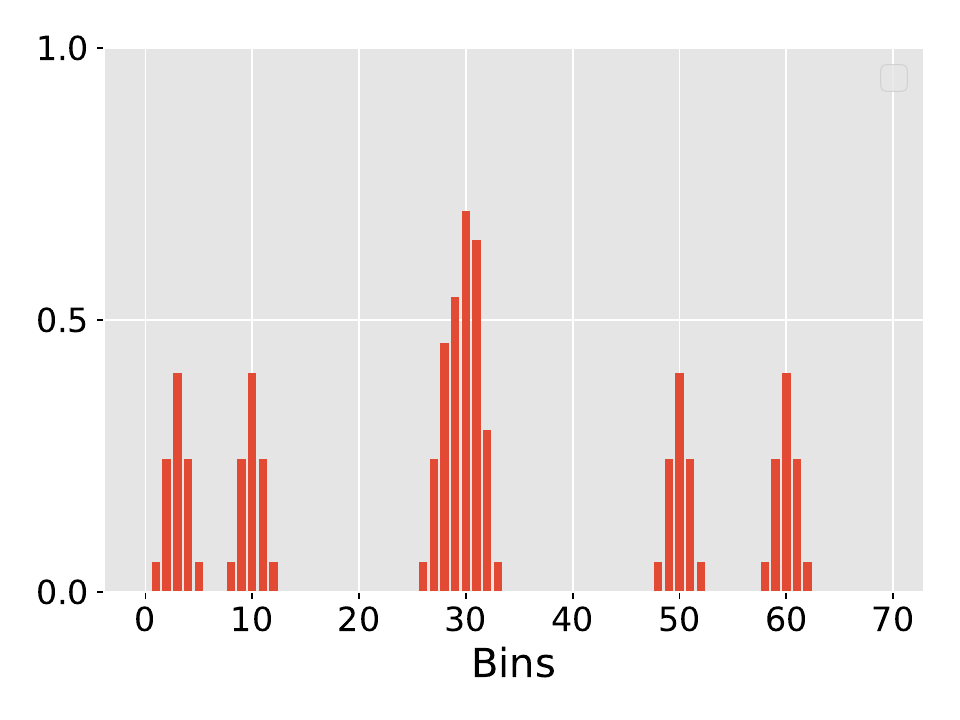}
}
\caption{Matched Filtering Output}
\end{subfigure}
\vspace{-5pt}
\caption{An example of a raw histogram and matched filtering output for a LiDAR measurement under extremely low SBR.}
\label{fig:toyex}
\end{figure}

We use matched filtering output from the timing histograms to find the peak and compute the probability attribute for each point. Under extreme conditions of low signal, there may not be any bin with more than one detected photon. The matched filtering output can still provide a dominant peak corresponding to the signal, and hence a correct point in the point cloud, if there are `clusters' of photon detections in nearby bins. Figure \ref{fig:toyex} illustrates this using a histogram and its matched filtering output.

\begin{figure}[ht!]
\centering
\begin{subfigure}[justification=centering]{0.48\linewidth}
\frame{
\includegraphics[width=\linewidth]{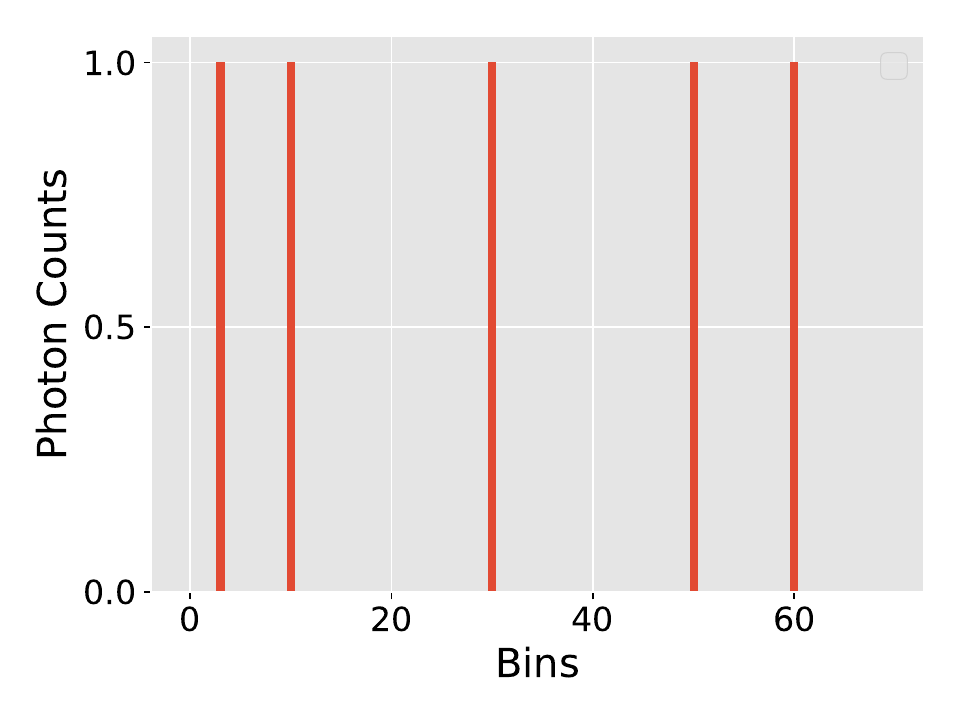}
}
\caption{Raw Histogram}
\end{subfigure}
\begin{subfigure}[justification=centering]{0.48\linewidth}
\frame{
\includegraphics[width=\linewidth]{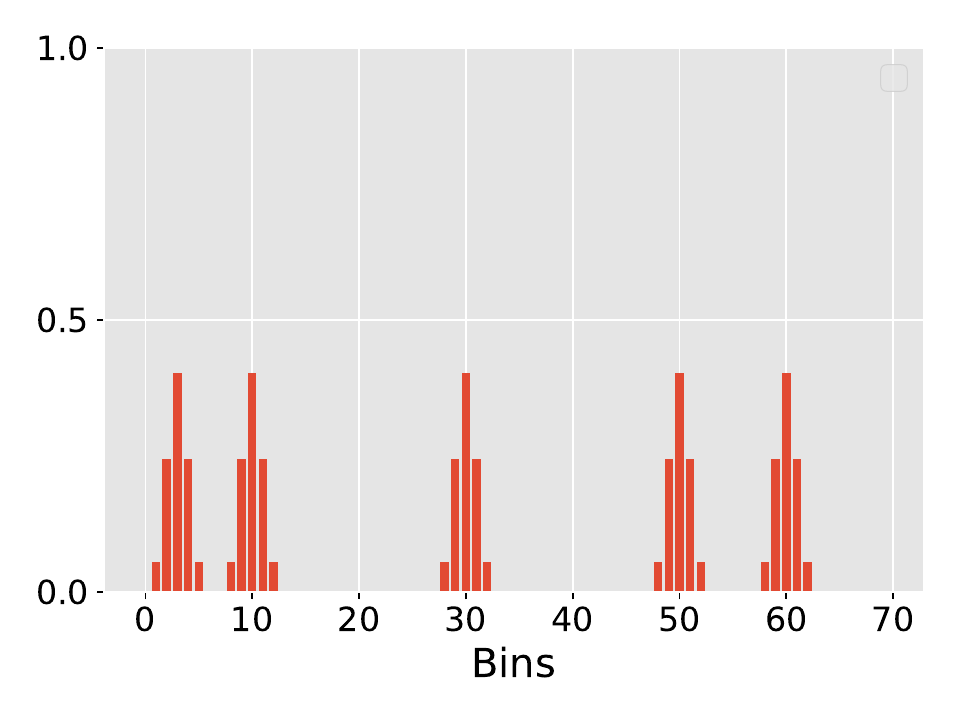}
}
\caption{Matched Filtering Output}
\end{subfigure}
\vspace{-5pt}
\caption{An example of a raw histogram and matched filtering output for a LiDAR measurement under extremely low SBR.}
\label{fig:toyex_sparse}
\end{figure}

Under ultra-low SBR, or when there is no return signal, there could be multiple sparsely located bins with single photon detection. Figure \ref{fig:toyex_sparse} shows an example to illustrate this scenario. In such cases, there is no peak corresponding to more than one photon in the matched filtering output. We use a minimum height threshold to ignore such points in the 3D point cloud.

\end{document}